\documentclass{article}

\usepackage{microtype}
\usepackage{graphicx}
\usepackage{booktabs} % for professional tables
\usepackage{multirow}

\usepackage{hyperref}

\usepackage[accepted]{icml2019}

\usepackage{subcaption} 
\usepackage{amsmath}
\usepackage{amssymb}

\usepackage{amsthm}
\newtheorem{theorem}{Theorem}
\newtheorem{lemma}{Lemma}

\newcommand{\Set}[1]{\mathbb{#1}}
\newcommand{\emb}{z}
\newcommand{\Emb}{Z}
\newcommand{\EmbS}{\bar{Z}}
\newcommand{\Am}{\hat{A}}
\newcommand{\loss}{\mathcal{L}}
\newcommand{\norm}[1]{\vert\vert #1 \vert\vert_0}

\newcommand{\lm}{\hat{M}}
\newcommand{\n}{N}
\newcommand{\attrand}{$\mathcal{B}_{rnd}~$}
\newcommand{\atteig}{$\mathcal{B}_{eig}~$}
\newcommand{\attdeg}{$\mathcal{B}_{deg}~$}
\newcommand{\attgrad}{$\mathcal{A}_{DW_2}~$}
\newcommand{\attpert}{$\mathcal{A}_{DW_3}~$}
\newcommand{\atttar}{$\mathcal{A}_{class}~$}
\newcommand{\attlnk}{$\mathcal{A}_{link}~$}
\newcommand{\ltarget}{$\mathcal{T}$}

\newcommand*\rot{\rotatebox{90}}

\icmltitlerunning{Adversarial Attacks on Node Embeddings via Graph Poisoning}

\begin{document}

\twocolumn[
\icmltitle{Adversarial Attacks on Node Embeddings via Graph Poisoning}

% It is OKAY to include author information, even for blind
% submissions: the style file will automatically remove it for you
% unless you've provided the [accepted] option to the icml2019
% package.

% List of affiliations: The first argument should be a (short)
% identifier you will use later to specify author affiliations
% Academic affiliations should list Department, University, City, Region, Country
% Industry affiliations should list Company, City, Region, Country

% You can specify symbols, otherwise they are numbered in order.
% Ideally, you should not use this facility. Affiliations will be numbered
% in order of appearance and this is the preferred way.
%\icmlsetsymbol{equal}{*}

\begin{icmlauthorlist}
\icmlauthor{Aleksandar Bojchevski}{to}
\icmlauthor{Stephan G\"unnemann}{to}
\end{icmlauthorlist}

\icmlaffiliation{to}{Technical University of Munich, Germany}

\icmlcorrespondingauthor{Aleksandar Bojchvski}{a.bojchevski@in.tum.de}

% You may provide any keywords that you
% find helpful for describing your paper; these are used to populate
% the "keywords" metadata in the PDF but will not be shown in the document
\icmlkeywords{Machine Learning, ICML}

\vskip 0.3in
]

% this must go after the closing bracket ] following \twocolumn[ ...

% This command actually creates the footnote in the first column
% listing the affiliations and the copyright notice.
% The command takes one argument, which is text to display at the start of the footnote.
% The \icmlEqualContribution command is standard text for equal contribution.
% Remove it (just {}) if you do not need this facility.

\printAffiliationsAndNotice{}  % leave blank if no need to mention equal contribution
%\printAffiliationsAndNotice{\icmlEqualContribution} % otherwise use the standard text.

\begin{abstract}
The goal of network representation learning is to learn low-dimensional node embeddings that capture the graph structure and are useful for solving downstream tasks.	
However, despite the proliferation of such methods, there is currently no study of their robustness to adversarial attacks.
We provide the first adversarial vulnerability analysis on the widely used family of methods based on random walks.
We derive efficient adversarial perturbations that poison the network structure and have a negative effect on both the quality of the embeddings and the downstream tasks.
We further show that our attacks are transferable since they generalize to many models and are successful even when the attacker is restricted.
\end{abstract}

\section{Introduction}
Unsupervised node embedding (network representation learning) approaches are becoming increasingly popular and achieve state-of-the-art performance on many network learning tasks \cite{cai2017comprehensive}.  
The goal is to embed each node in a low-dimensional feature space such that the graph's structure is captured. The learned embeddings are subsequently used for downstream tasks such as link prediction, node classification, community detection, and visualization.
Among the variety of proposed approaches, techniques based on random walks (RWs) \cite{perozzi2014deepwalk, grover2016node2vec} are often employed since they incorporate higher-order relational information.
Given the increasing popularity of these methods there is a strong need for an analysis of their robustness.
In particular, we aim to study the existence and effects of \textit{adversarial perturbations}. A large body of research shows that both traditional and deep learning methods can easily be fooled/attacked: even slight deliberate data perturbations can lead to wrong results \cite{GoodfellowSS14,Mei2015Machine.pdf,1608.04644.pdf,1704.08006.pdf,Houdini.pdf,1703.06748.pdf,abs-1712-05526}.

While adversarial attacks for graph models have been proposed recently \cite{zugner2018adversarial, dai2018adversarial, zugner2019adversarial}, they are all limited to the semi-supervised learning setting. In contrast, this is the first work that studies adversarial perturbations for \emph{unsupervised} embeddings. This is critical, since especially in domains where graph embeddings are used (e.g.\ the web) adversaries are common and false data is \emph{easy to inject}: e.g. spammers can easily create fake followers on social networks.
Can we construct attacks that do no rely on a specific downstream task? Are node embedding methods just as easily fooled, since compared to semi-supervised models they do not incorporate a supervision signal that can be exploited?

Barring the few above-mentioned attacks on graphs most existing works on adversarial attacks perturb the features of individual instances. In our case however, since we are operating on plain graph data (no features are available) we perturb the interactions (edges) between instances instead. Manipulating the network structure (the graph) is a common scenario with link spam farms \cite{gyongyi2005link} and Sybil attacks \cite{yu2006sybilguard} as typical examples. 

Moreover, since node embeddings are typically trained in an unsupervised and transductive fashion we cannot rely on a single end-task that our attack might exploit to find appropriate perturbations, and we have to handle a challenging poisoning attack where the model is learned \textit{after} the attack. That is, the model cannot be assumed to be static as in most existing attacks.
Lastly, since graphs are discrete, gradient-based approaches \cite{li2016data,Mei2015Machine.pdf} for finding adversarial  perturbations that were designed for continuous data are not well suited. 
In particular, for methods based on random walks the gradient computation is not directly possible since sampling random walks is not a differentiable operation.
The question is how to design efficient algorithms that are able to find adversarial perturbations in such a challenging -- discrete and combinatorial -- graph~domain? 

We propose a principled strategy for adversarial attacks on unsupervised node embeddings. Exploiting results from eigenvalue perturbation theory \cite{stewart1990matrix} we are able to efficiently solve a challenging bi-level optimization problem associated with the poisoning attack.
We assume an attacker with full knowledge about the data and the model, thus ensuring reliable vulnerability analysis in the worst case. Nonetheless, our experiments on transferability demonstrate that our strategy generalizes -- attacks learned based on one model successfully fool other models as well. Additionally, we study the effect of restricting the attacker.

Overall, we shed light on an important problem that has not been studied so far. We show that node embeddings are sensitive to adversarial attacks. Relatively few changes are needed to significantly damage the quality of the embeddings even in the scenario where the attacker is restricted. Furthermore, our paper highlights that more work is needed to make node embeddings robust to adversarial perturbations and thus readily applicable in production systems.
\section{Related Work}
We focus on unsupervised node embedding approaches based on random walks (RWs) and further show how one can easily apply a similar analysis to attack spectral-based node embeddings.
For a recent survey, also of other non-RW based approaches, we refer to \citet{cai2017comprehensive}. Moreover, while many semi-supervised learning methods \cite{defferrard2016convolutional,kipf2016semi,klicpera2018combining} have been introduced, we focus on unsupervised methods since they are often used in practice due to their flexibility in simultaneously solving various downstream tasks.

\textbf{Adversarial attacks.}
Attacking machine learning models has a long history, with seminal works on SVMs and logistic regression \cite{1206.6389.pdf,Mei2015Machine.pdf}.
Neural networks were also shown to be highly sensitive to small adversarial perturbations to the input \cite{szegedy2013intriguing,GoodfellowSS14}. 
While most works focus on image classification, recent works also study adversarial examples in other domains \cite{grosse2017adversarial,1704.08006.pdf}.

Different taxonomies exist characterizing the adversaries based on their goals, knowledge, and capabilities
\cite{1511.07528.pdf,biggio2014security,SuperVorlage.pdf}.
The two dominant attacks types are poisoning attacks targeting the training data (the model is trained \textit{after} the attack) and evasion attacks targeting the test data/application phase (the learned model is assumed fixed). 
Compared to evasion attacks, poisoning attacks are far less studied
\cite{Mei2015Machine.pdf,li2016data,koh2017understanding,SuperVorlage.pdf,abs-1712-05526} since they usually require solving a challenging bi-level optimization problem.

\textbf{Attacks on semi-supervised graph models.}
The robustness of semi-supervised graph classification methods to adversarial attacks has recently been analyzed \cite{zugner2018adversarial,dai2018adversarial,zugner2019adversarial}. The first work, introduced by \citet{zugner2018adversarial}, linearizes a graph convolutional network (GCN) \cite{kipf2016semi}  to derive a closed-form expression for the change in class probabilities for a given edge/feature perturbation. They calculate a score for each possible edge flip based on the classification margin and greedily pick the top edge flips with highest scores. Later, \citet{dai2018adversarial} proposed a reinforcement (Q-)learning formulation where they decompose the selection of relevant edge flips into selecting the two end-points. \citet{zugner2019adversarial} develop a general attack on the training procedure of GCN using meta-gradients.
All three approaches focus on the semi-supervised graph classification task and take advantage of the supervision signal to construct the attacks. In contrast, our work focuses on general attacks on \emph{unsupervised} node embeddings applicable to many downstream tasks.

\textbf{Manipulating graphs.}
There is an extensive literature on optimizing the graph structure to manipulate: information spread in a network \cite{khalil2014scalable,chen2016eigen}, user opinions \cite{chaoji2012recommendations,amelkin2017disabling}, shortest paths \cite{phillips1993network,israeli2002shortest}, page rank scores \cite{csaji2014pagerank}, and other metrics \cite{chan2014make}.
In the context of graph clustering, \citet{chen2017practical} measure the performance changes when injecting noise to a bi-partite graph of DNS queries, but do not focus on automatically generating attacks. 
\citet{zhao2018data} study poisoning attacks on multi-task relationship learning, although they exploit relations between tasks, they still deal with the classic scenario of i.i.d. instances within each task.

\textbf{Robustness and adversarial training.}
Robustification of machine learning models, including graph based models \cite{bojchevski2017robust, zugner2019certifiable}, has been studied and is known as adversarial/robust machine learning. These approaches are out of the scope for this paper.
Adversarial training, e.g.\ via GANs \cite{dai2017adversarial}, is similarly beyond our scope since the goal is to improve the embeddings, while our goal is to asses the vulnerability of existing embedding methods to adversarial perturbations.
\section{Attacking Node Embeddings}
We study poisoning attacks on the graph structure -- the attacker is capable of adding or removing (flipping) edges in the original graph within a given budget. We focus mainly on approaches based on random walks and extend the analysis to spectral approaches (Sec.\ \ref{sec:spectral} in the appendix).

\subsection{Background and Preliminaries}
Let $G = (V, E)$ be an undirected unweighted graph where $V$ is the set of nodes, $E$ is the set of edges, and $A \in \{0,1\}^{|V| \times |V|}$ is the adjacency matrix. 
The goal of network representation learning is to find a low-dimensional embedding $\emb_v \in \Set{R}^K$ for each node with $K \ll |V|$.
This dense low-dimensional representation should preserve information about the network structure -- nodes similar in the original network should be close in the embedding space.
DeepWalk \cite{perozzi2014deepwalk} and node2vec \cite{grover2016node2vec} learn an embedding based on RWs by adapting the skip-gram architecture \cite{mikolov2013efficient} for learning word embeddings. 
They sample finite (biased) RWs and use the co-occurrence of node-context pairs in a given window in each RW as a measure of similarity.
To learn $\emb_v$ they maximize the probability of observing $v$'s neighborhood.

\subsection{Attack Model}
We denote with $\Am$ the adjacency matrix of the graph obtained after the attacker has modified certain entries in $A$.
We assume the attacker has a given, fixed budget and is only capable of modifying $f$ entries, i.e. $||\Am - A||_0 = 2f$ (times $2$ since $G$ is undirected).
The goal of the attacker is to damage the quality of the learned embeddings, which in turn harms subsequent learning tasks such as node classification or link prediction that use the embeddings. We consider both a general attack that aims to degrade the embeddings of the network as a whole, and a targeted attack that aims to damage the embeddings regarding a specific target~/~task.

The quality of the embeddings is measured by the loss $\loss(A, \Emb)$ of the model under attack, with lower loss corresponding to higher quality, where $\Emb \in \Set{R}^{N \times K}$ is the matrix containing the embeddings of all nodes. Thus, the goal of the attacker is to \textit{maximize} the loss. 
We can formalize this as the following bi-level optimization problem:
\begin{align}
	&\Am^* = \arg\max_{\Am \in \{0,1\}^{N\times N} } \loss(\Am, Z^*) \nonumber
	&Z^* = \min_{Z} \loss(\Am, Z)  \\ \nonumber
	&\text{subj. to} ~  \norm{\Am-A} = 2f \text{~,~} \Am=\Am^T 
\end{align}
Here, $Z^*$ is always the \emph{'optimal'} embedding resulting from the (to be optimized) graph $\Am$, i.e.\ it minimizes the loss, while the attacker tries to maximize the loss. Solving such a problem is challenging given its discrete and combinatorial nature, thus we derive efficient approximations.

\subsection{General Attack}
The first step in RW-based embedding approaches is to sample a set of random walks that serve as a training corpus further complicating the bi-level optimization problem.
We have $Z^* = \min_{Z} \loss(\{r_1, r_2, \dots \}, Z)$ with $r_i \sim RW(\Am)$, where $RW$ is an intermediate stochastic procedure that generates RWs given the graph $\Am$ which we are optimizing. 
By flipping (even a few) edges in the graph, the attacker necessarily changes the set of possible RWs, thus changing the training corpus.
Therefore, this sampling procedure precludes any gradient-based methods. 
To tackle this challenge we leverage recent results that show that (given certain assumptions) RW-based embedding approaches are implicitly factorizing the Pointwise Mutual Information (PMI) matrix \cite{yang2015comprehend, qiu2017network}.
We study DeepWalk as an RW-based representative approach since it's one of the most popular methods and has many extensions.
Specifically, we use the results from \citet{qiu2017network} to sidestep the stochasticity induced by sampling random walks.

\begin{lemma}[\citet{qiu2017network}]
	\label{lem:qiu}
	DeepWalk is equivalent to factorizing $\lm = \log(\max(M, 1))$ with 
	\begin{equation}\label{eq:S}
		M = \tfrac{vol(A)}{T\cdot b}~S
		,\quad
		S=\big(\sum_{r=1}^{T} P^r \big) D^{-1}
		,\quad
		P=D^{-1}A
	\end{equation}
	 where the embedding $Z^*$ is obtained by the Singular Value Decomposition (SVD) of $\lm=U\Sigma V^T$ using the top-$K$ largest singular values~/~vectors, i.e. $Z^*= U_K\Sigma_K^{1/2}$. 
\end{lemma}
Here, $D$ is the diagonal degree matrix with $D_{ii} = \sum_{j}A_{ij}$, $T$ is the window size, $b$ is the number of negative samples and $vol(A) = \sum_{i,j} A_{ij}$ is the volume. Since $M$ is sparse and has many zero entries the matrix $\log(M)$ where the $\log$ is elementwise is ill-defined and dense. To cope with this, similar to the Shifted Positive PMI (PPMI), approach the elementwise maximum is introduced to form $\lm$. 
Using this insight we see that DeepWalk is equivalent to optimizing $\min_{\lm_K}||\lm - \lm_K||_F^2$ where $\lm_K$ is the best rank-$K$ approximation to $\lm$.
This in turn means that the loss for DeepWalk when using the \textit{optimal} embedding $Z^*$ for a given graph $A$ is $\loss_{DW_1}(A, Z^*) = \big[\sum_{p=K+1}^{|V|} \sigma_p ^ 2\big]^{1/2}$ where $\sigma_p$ are the singular values of $\lm(A)$ sorted decreasingly $\sigma_1\ge \sigma_2 \dots\ge \sigma_{|V|}$. This result shows that we do not need to construct random walks, nor do we have to (explicitly) learn the embedding $Z^*$ -- it is implicitly considered via the singular values of $\lm(A)$. Accordingly, we have transformed the bi-level problem into a single-level optimization problem. However, maximizing $\loss_{DW_1}$ is still challenging due to the SVD and the discrete nature of the problem.

\textbf{Gradient based approach.}
Maximizing $\loss_{DW_1}$ with a gradient-based approach is not straightforward since we cannot easily backpropagate through the SVD. To tackle this challenge we exploit ideas from eigenvalue perturbation theory \citep{stewart1990matrix} to efficiently approximate $\loss_{DW_1}(A)$ in closed-form without needing to recompute the SVD.

\begin{theorem}
	\label{theo:naive_grad}
	Let $A$ be the initial adjacency matrix and $\lm(A)$ be the respective co-occurrence matrix. Let $u_p$ be the $p$-th eigenvector corresponding to the $p$-th largest eigenvalue of $\lm$.
	Given a perturbed matrix $A'$, with $A'=A+\Delta A$, and the respective change $\Delta \lm$, 
	$\loss_{DW_1}(A') \approx \big[\sum_{p=K+1}^{\n} \big(u_p^T (\lm + \Delta \lm) u_p\big)^2\big]^{1/2} =: \loss_{DW_2}(A') \nonumber
	$ is an approximation of the loss and the error is bounded by $|\loss_{DW_1}(A') - \loss_{DW_2}(A') | \le ||\Delta \lm||_F$.
\end{theorem}

The proof is given in the appendix.
For a small $\Delta A$ and thus small $\Delta \lm$ we obtain a very good approximation, and if $\Delta A = \Delta \lm = 0$ then the loss is exact.
Intuitively, we can think of using eigenvalue perturbation as analogous to taking the gradient of the loss w.r.t. $\lm(A)$. Now, gradient-based optimization is efficient since $\nabla_A \loss_{DW_2}(A)$ avoids recomputing the eigenvalue decomposition.
The gradient provides useful information for a small $\epsilon$ change, however, here we are considering discrete flips, i.e. $\epsilon=\pm1$ so its usefulness is limited.
Furthermore, using gradient-based optimization requires a dense instantiation of the adjacency matrix which has complexity $O(|V|^2)$ in both runtime and memory, infeasible for large graphs. This motivates the need for our more advanced approach.

\textbf{Sparse closed-form approach.}
Our goal is to efficiently compute the change in the loss $ \loss_{DW_1}(A)$ given a set of flipped edges. To do so we will analyze the change in the spectrum of some of the intermediate matrices and then derive a bound on the change in the spectrum of the co-occurrence matrix, which in turn will give an estimate of the loss. First, we need some results.

\begin{lemma}
	\label{lemma:Seq}
	The matrix $S$ in Eq.\ \ref{eq:S} is equal to 
	$
	S  = U (\sum_{r=1}^{T} \Lambda^r) U^T
	$
	where the matrices $U$ and $\Lambda$ contain the eigenvectors and eigenvalues solving the generalized eigen-problem $Au=\lambda Du$.
\end{lemma}
The proof is given in the appendix.
We see that the spectrum of $S$ (and the spectrum of $M$ by taking the scalars into account) is obtainable from the generalized spectrum of $A$.  
In contrast to Lemma \ref{lemma:Seq}, \citet{qiu2017network} factorize $S$ using the (non-generalized) spectrum of $A_{norm}:=D^{-1/2} A D^{-1/2}$. As we will show, our formulation using the generalized spectrum of $A$ is key for an efficient approximation.

Let $A'=A+\Delta A$ be the adjacency matrix after the attacker performed some edge flips. As above,  by computing the generalized spectrum of $A'$, we can estimate the spectrum of the resulting $S'$ and $M'$. However, recomputing the eigenvalues $\lambda'$ of $A'$ for every possible set of edge flips is still not efficient for large graphs, preventing an effective application of the method. Thus, we derive our first main result: an efficient approximation bounding the change in the singular values of $M'$ for any edge flip.

\begin{theorem}
	\label{theo:delta_lambda}
	Let $\Delta A$ be a matrix with only 2 non-zero elements, namely $\Delta A_{ij} = \Delta A_{ji} = 1 - 2A_{ij}$ corresponding to a single edge flip $(i,j)$, and $\Delta D$ the respective change in the degree matrix, i.e. $A'=A+\Delta A$ and $D'=D+\Delta D$.
	Let  $u_{y}$ be the $y$-th generalized eigenvector of $A$ with generalized eigenvalue $\lambda_y$.
	Then the generalized eigenvalue $\lambda'_y$ of $A'$ solving $A'u'_y = \lambda'_y D'u'_y$ is approximately $\lambda'_y \approx \lambda_y + \Delta \lambda_y := \tilde{\lambda'_y} $ with:
	\begin{equation}\label{eq:perturb}
	\Delta \lambda_y=\Delta w_{ij}(2u_{yi}\cdot u_{yj} - \lambda_y(u_{yi}^2 + u_{yj}^2 ))
	\end{equation}
	where $u_{yi}$ is the $i$-th entry of the vector $u_y$, and $\Delta w_{ij} = (1-2A_{ij})$ indicates the edge flip, i.e $\pm1$.
\end{theorem}
The proof is given in the appendix. By working with the generalized eigenvalue problem in Theorem \ref{theo:delta_lambda} we were able to express $A'$ and $D'$ after flipping an edge as \emph{additive} changes to $A$ and $D$, this in turn enabled us to leverage results from eigenvalue perturbation theory to efficiently approximate the change in the spectrum. If we used $A_{norm}$ instead, the change to $A'_{norm}$ would be multiplicative hindering efficient approximation.
Using Eq.\ \ref{eq:perturb}, instead of recomputing $\lambda'$ we only need to compute $\Delta \lambda$ to obtain the approximation $\tilde{\lambda'}$
significantly reducing the complexity when evaluating different edge flips $(i,j)$. 
Using this result, we can now efficiently bound the change in the singular values of $S'$.
\begin{lemma}
	\label{lemma:cor_loss}
	Let $A'$ be defined as before and $S'$ be the resulting matrix.
	The singular values of $S'$ are bounded:
	$
	\label{eq:estimate}
	\sigma_p(S') \le \tilde{\sigma}_p:=
	\tfrac{1}{{d'_{min}}}\cdot 
	\big\vert \sum_{r=1}^{T} (\tilde{\lambda'}_{\pi(p)}) ^r \big\vert
	$ where $\pi$ is a permutation simply ensuring that the final $\tilde{\sigma}_p$ are sorted decreasingly, and $d'_{min}$ is the smallest degree in $A'$.
\end{lemma}
We provide the proof in the appendix. 
Now we can efficiently compute the loss for a rank-$K$ factorization of $M'$, which we would obtain when performing the edge flip $(i,j)$, i.e. 
$
\mathcal{L}_{DW_3}(A')= \frac{vol(A) + 2\Delta w_{ij}}{T\cdot b}\big[\sum_{p=K+1}^{|V|} \tilde{\sigma}_p^2\big]^{1/2}
$, where $\tilde{\sigma}_p$ is obtained by applying Lemma \ref{lemma:cor_loss} and Theorem \ref{theo:delta_lambda} and the leading constants follow from Lemma \ref{lem:qiu}.

While the original loss $\mathcal{L}_{DW_1}$  is based on the matrix $\lm=\log(\max(M, 1))$, there are unfortunately currently no tools available to analyze the (change in the) spectrum of $\lm$ given the spectrum of $M$. Therefore, we use  $\mathcal{L}_{DW_3}$ as a surrogate loss for $\mathcal{L}_{DW_1}$ (\citet{yang2015network} similarly exclude the element-wise logarithm). As our experiments show, the surrogate loss is effective and we can successfully attack the node embeddings that factorize the actual co-occurrence matrix $\lm$, as well as the original skip-gram model. Similarly, spectral embedding methods \cite{von2007tutorial}, factorize the graph Laplacian and have a strong connection to the RW based approaches. We provide an analysis of their adversarial vulnerability in the appendix (Sec.\ \ref{sec:spectral}).

\textbf{The overall algorithm.}
Our goal is to maximize $\mathcal{L}_{DW_3}$ by performing $f$ edge flips. 
While Eq.\ \ref{eq:perturb} enables us to efficiently compute the loss for a single edge, there are still $\mathcal{O}(|V|^2)$ possible flips.
To reduce the complexity when~\emph{adding} edges we instead form a candidate set by randomly sampling $C$ candidate pairs (non-edges). This introduces a further approximation that nonetheless works well in practice. Since real graphs are usually sparse, for \emph{removing}, all edges are viable candidates with one random edge set aside for each node to ensure we do not have singleton nodes.
For every candidate we compute its impact on the loss via $\mathcal{L}_{DW_3}$ and greedily choose the top $f$ flips.\footnote{Periodically recomputing the exact eigenvalues when using the greedy approach did not show any benefits. Code and data available at \href{https://www.kdd.in.tum.de/node_embedding_attack}{https://www.kdd.in.tum.de/node\_embedding\_attack}.}

The runtime complexity of our overall approach is then: $\mathcal{O}(|V|\cdot |E|+C\cdot |V|\log |V| ) $.
First, we can compute the generalized eigenvectors of $A$ in a sparse fashion in $\mathcal{O}(|V|\cdot |E|)$. Then we sample $C$ candidate edges, and for each we can compute the approximate eigenvalues in constant time (Theorem \ref{theo:delta_lambda}). To obtain the final loss, we sort the values leading to the overall complexity.
For the examined datasets the wall-clock time for our approach is negligible: on the order of few seconds when calculating the change in eigenvalues.
Furthermore, our approach is trivially parallelizable since every candidate edge flip can be evaluated in parallel.

\vspace*{-0.35em}\subsection{Targeted Attack}
\label{sec:targeted_attack}
If the goal of the attacker is to attack a specific target node $t \in V$, or a specific downstream task, it is suboptimal to maximize the overall loss via $\mathcal{L}_{DW_*}$. Rather, we should define some other \emph{target specific} loss that depends on $t$'s embedding -- replacing the loss function of the \emph{outer} optimization by another one operating on $t$'s embedding.
Thus, for any edge flip $(i, j)$ we now need the change in $t$'s embedding -- meaning changes in the eigen\emph{vectors} -- which is inherently more difficult to compute compared to changes in eigen/singular-\emph{values}. We study two cases: misclassifying a target node (i.e. node classification tasks) and manipulating the similarity of node pairs (i.e. link prediction task).

\textbf{Surrogate embeddings.}
We define surrogate embeddings such that we can efficiently estimate the change for a given edge flip. Specifically, instead of performing an SVD of $M$ (or equivalently $S$ scaled) we define $\EmbS^*=U (\sum_{r=1}^{T} \Lambda^r)$,
as in Lemma \ref{lemma:Seq}.
Experimentally, using $\EmbS^*$ instead of $\Emb^*$ as the embedding showed no significant change in the performance on downstream tasks. 
While we use these surrogate embeddings to select the adversarial edges, during evaluation we use the standard embeddings produced by DeepWalk.
Now, by approximating the generalized eigenvectors of $A'$, we can also approximate $\EmbS^*(A')$ in closed-form: 
\begin{theorem}
	\label{theo:delta_u}
	Let $\Delta A, \Delta D$ and $\Delta w_{ij}$ be defined as before, and $\Delta \lambda_y$ be the change in the y-th generalized eigenvalue $\lambda_y$ as derived in Theorem \ref{theo:delta_lambda}. Then, the y-th generalized eigenvector $u'_y$ of $A'$ after performing the edge flip $(i, j)$ can be approximated with:
	\begin{align}
\label{eq:delta_u}
		u'_y \approx u_y
		 - \Delta w_{ij} (A-\lambda_y D)^+ \big(-\Delta \lambda_y u_y \circ d \\ \nonumber
		+E_i(u_{yj} - \lambda_y u_{yi})
		+E_j(u_{yi} - \lambda_y u_{yj})
		 \big)
	\end{align}
	where $E_i(x)$ returns a vector of zeros except at position $i$ where the value is $x$, $d$ is a vector of the node degrees, $\circ$ is the Hadamard product, and $(\cdot)^+$ is the pseudo-inverse.
\end{theorem}
We provide the proof in the appendix. Computing Eq.\ \ref{eq:delta_u} seems expensive at first due to the pseudo-inverse term. However, note that this term does not depend on the particular edge flip we perform. Thus, we can pre-compute it once and furthermore, parallelize the computation for each $y$.
The additional complexity of computing the pseudo-inverse for all $y$ is $\mathcal{O}(K \cdot |V|^{2.373})$.
Similarly, we can pre-compute $u_{y}\circ d$, while the rest of the terms are all computable in $O(1)$.
Overall, the wall-clock time for computing the change in the eigenvectors is on the order of few minutes.
For any edge flip we can now efficiently compute the optimal embedding $\EmbS^*(A')$ using Eqs.\ \ref{eq:perturb} and \ref{eq:delta_u}. The t-th row of $\EmbS^*(A')$ is the desired embedding for a target node $t$ after the attack.

\textbf{Targeting node classification.}
\label{sec:targeted_class}
Our goal is to missclassify a target node $t$ given a downstream node classification task. To specify the targeted attack we need to define the candidate flips and the target-specific loss responsible for scoring the candidates. We let the candidate set contain all edges (and non-edges) directly incident to the target node, i.e. $\{(v, t) \vert v \ne t\}$.
We restricted our experiments to such candidate flips since initial experiments showed that they can do significantly more damage compared to candidate flips in other parts of the graph. This intuitively makes sense since the further away we are from node $t$ we can exert less influence on it. \citet{zugner2018adversarial} show similar results (e.g.\ see their indirect attack). Note that for the general (non-targeted) attack all edges/non-edges are viable candidates.

To obtain the loss, we first pre-train a classifier on the clean embedding $\EmbS^*$. Then we predict the class probabilities $p_t$ of the target $t$ using the compromised $\EmbS^*_{t, \cdot}$ estimated for a given candidate flip and we calculate the classification margin $m(t) = p_{t,c(t)} -  \max_{c \ne c(t)} p_{t,c}$, where $c(t)$ is the ground-truth class for $t$. That is, our loss is the difference between the probability of the ground truth and the next most probable class after the attack. Finally, we select the top $f$ flips with smallest margin $m$ (note when $m(t)<0$ node $t$ is misclassified).
In practice, we average over ten randomly trained logistic regression classifiers. In future work we plan to treat this as a tri-level optimization problem.

\textbf{Targeting link prediction.}
The goal of this targeted attack is given a set of target node pairs $\mathcal{T} \subset V \times V$ to decrease the similarity between the nodes that have an edge, and increase the similarity between nodes that do not have an edge by modifying \emph{other} parts of the graph (i.e.\ we disallow to directly flip pairs in \ltarget). For example, in an e-commerce graph representing users and items the goal might be to increase the similarity between a certain item and user by adding/removing connections between other users/items.
To achieve this goal, we first train an initial clean embedding on the graph excluding the edges in $\mathcal{T}$. Then, for a candidate flip we estimate the embedding $\EmbS^*$ (Eqs.\ \ref{eq:perturb} and \ref{eq:delta_u}) and use it to calculate the average precision score (AP score) on the target set \ltarget, with $\EmbS^*_{i} (\EmbS^*_{j})^T$ measuring the similarity of nodes $i$ and $j$
(i.e. the likelihood of the link $(i, j)$). Low AP score then indicates that the edges in $\mathcal{T}$ are less likely (non-edges more likely resp.). Finally, we pick the top $f$ flips with lowest AP scores and use them to poison the network.

\begin{figure*}[t!]
	\begin{subfigure}[h]{\linewidth}
		\centering
		\includegraphics[width=0.231\linewidth,clip,trim=0 0 270 0]{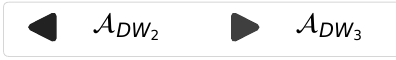}
		\hspace*{-3pt}\includegraphics[width=0.4\linewidth,clip,trim=170 0 0 0]{figs/legend}
	\end{subfigure}
	%	\begin{minipage}{.75\textwidth}
	\centering
	\begin{subfigure}[h]{0.33\linewidth}
		\includegraphics[width=\textwidth]{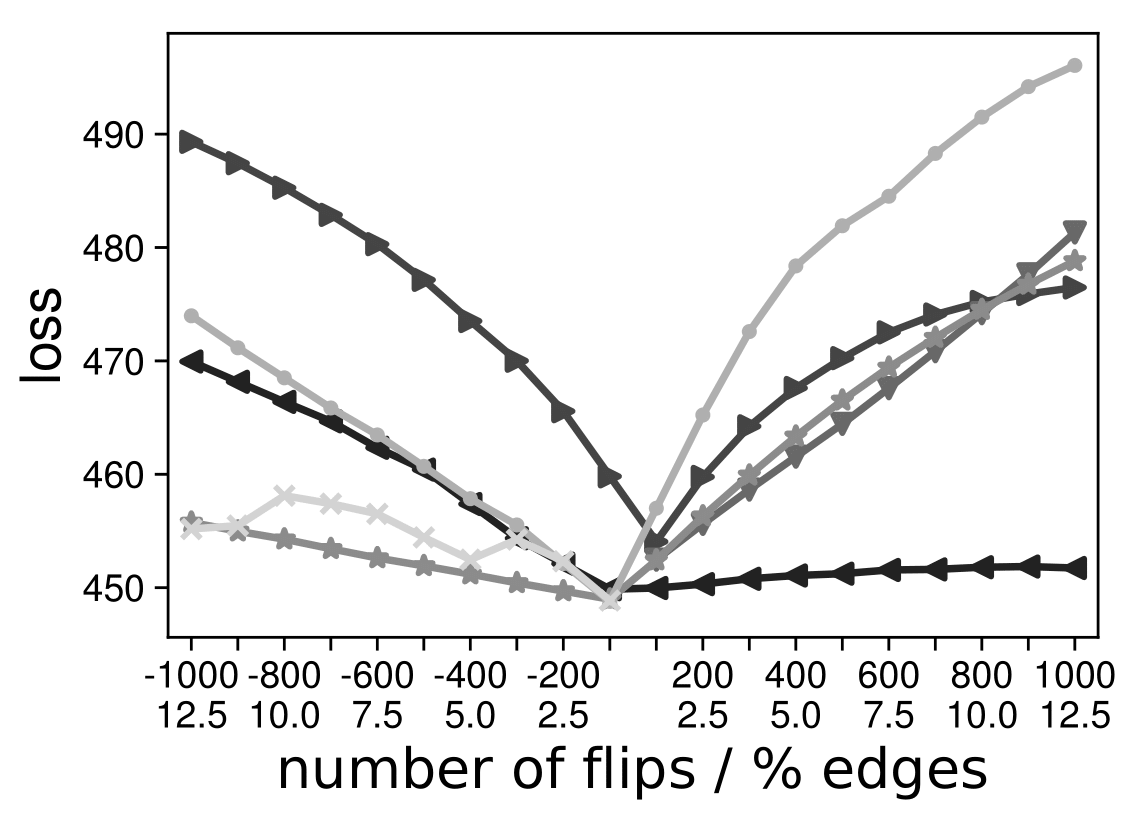}
		\caption{Cora: DeepWalk's loss}
		\label{fig_flips_loss}
	\end{subfigure}
	\begin{subfigure}[h]{0.33\linewidth}
		\includegraphics[width=\textwidth]{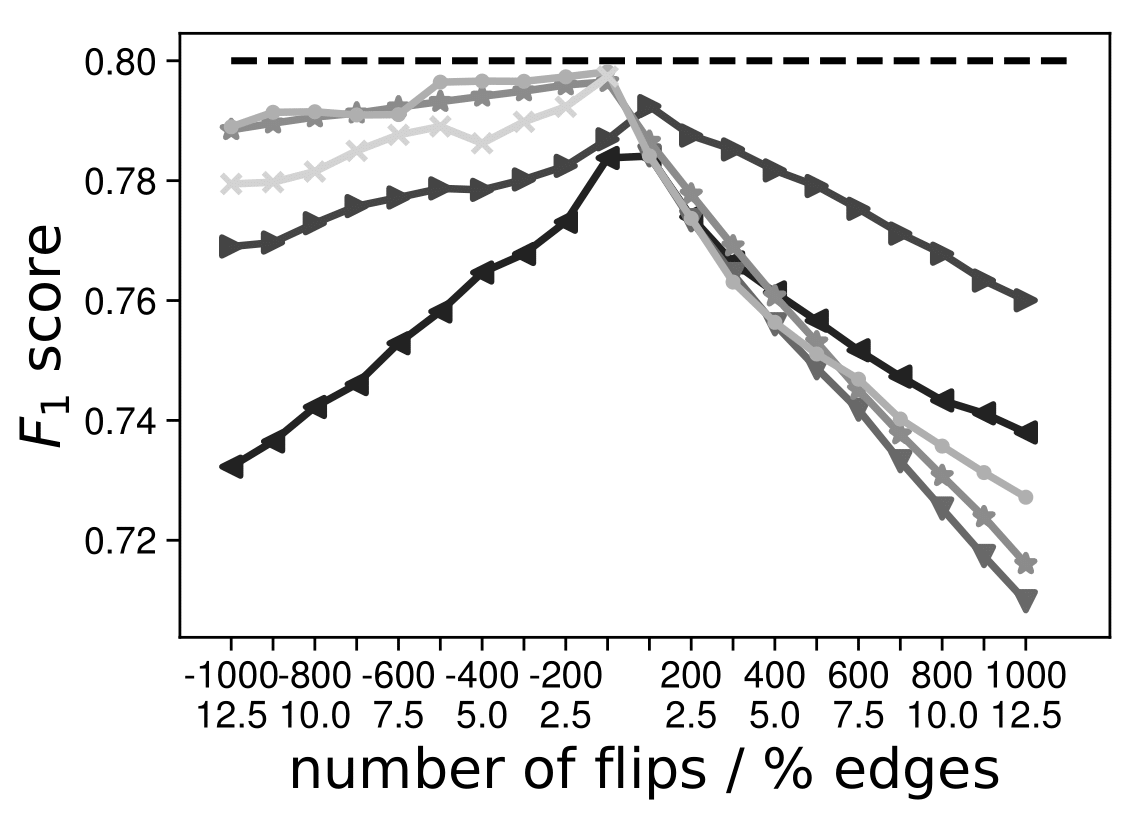}
		\caption{Cora: Classification}
		\label{fig_flips_f1}
	\end{subfigure}
	\begin{subfigure}[h]{0.33\linewidth}
		\includegraphics[width=\textwidth]{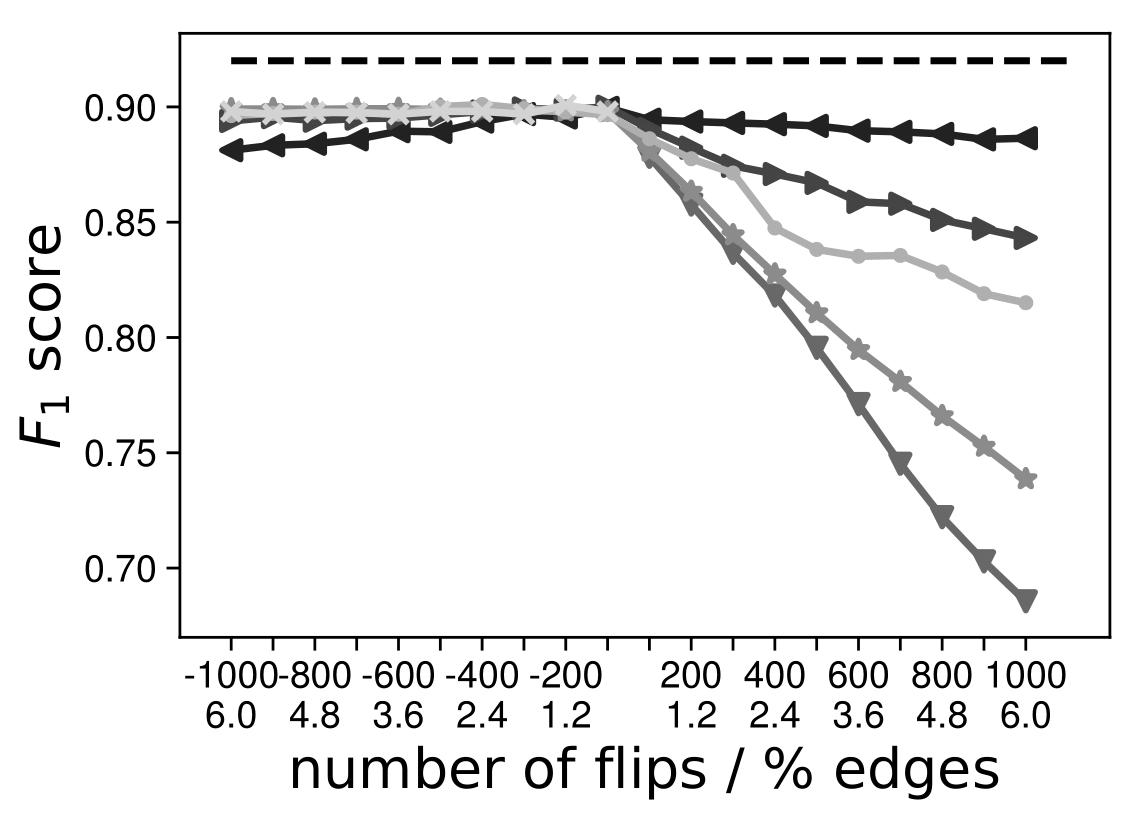}
		\caption{PolBlogs: Classification}
		\label{polblogs}
	\end{subfigure}
	\caption{Vulnerability of the embeddings under the general attack for increasing number of flips. Positive (resp. negative) numbers on the x-axis indicate adding (resp. removing) edges. The percentage of flips is w.r.t. the total number of edges in the clean graph. The dotted line shows the performance on the clean graph before attacking.}
	\label{fig:attacks_first}
\end{figure*}
\section{Experimental Evaluation}
Since this is the first work considering adversarial attacks on node embeddings there are no known baselines. Similar to methods that optimize the graph structure
\citep{chen2016eigen, khalil2014scalable}
we compare with several strong baselines. \attrand randomly flips edges (we report averages over ten seeds), \atteig removes edges based on their eigencentrality in the line graph $L(A)$, and \attdeg removes edges based on their degree centrality in $L(A)$ (equivalently sum of degrees in the graph).
When adding edges we use the same baselines as above now calculated on the complement graph except for \atteig since it is infeasible to compute even for medium size graphs. 
\attgrad denotes our gradient based attack, \attpert our closed-form attack, \attlnk our targeted link prediction attack, and \atttar is our targeted node classification attack.

The size of the sampled candidate set for adding edges under the general attack is 20K (we report averages over five trials).
To evaluate the targeted link prediction attack we form the target pairs $\mathcal{T}$ by randomly sampling $10\%$ of the edges from the clean graph and three times as many non-edges. This setup reflects the fact that in practice the attackers often care more about increasing the likelihood of a new edge, e.g. increasing the chance of recommending an item to a user. 

We aim to answer the following questions:
(Q1) how good are our approximations of the loss;
(Q2) how much damage is caused to the overall embedding quality by our attacks compared to the baselines;
(Q3) can we still perform a successful attack when the attacker is restricted;
(Q4) what characterizes the selected (top) adversarial edges;
(Q5) how do the targeted attacks affect downstream tasks; and (Q6) are the selected adversarial edges transferable to other models. 

We set DeepWalk's hyperparameters to: $T=5,b=5,K=64$ and use logistic regression for classification.
We analyze three datasets: Cora $(N=2810, \vert E\vert=15962$, \citet{mccallum2000automating, bojchevski2017deep}) and Citeseer $(N=2110,\vert E\vert=7336$, \citet{giles1998citeseer}) are citation networks commonly used to benchmark embedding approaches, and PolBlogs $(N=1222,\vert E\vert=33428$, \citet{adamic2005political}) is a graph of political blogs.
Since we are in the poisoning setting, in all experiments after choosing the top $f$ flips we re-train the standard embeddings produced by DeepWalk and report the final performance.
Note, for the \emph{general attack} the downstream node classification performance is \emph{only a proxy} for estimating the embedding quality after the attack, it is not our goal to damage this task, but rather to attack the unsupervised embeddings in general.

\subsection{Approximation Quality}
To estimate the approximation quality we randomly select 20K candidates from the Cora graph and we compute Pearson's $R$ score between the actual loss (including the elementwise logarithm) and our approximations. For example, for dimensionality $K=32$ we have $R(\loss_{DW_2}, \loss_{DW_1})=0.11$ and 	$R(\loss_{DW_3}, \loss_{DW_1})=0.90$ showing that our closed-form strategy approximates the loss significantly better than the gradient-based one. This also holds for $K=64, 128$.
Moreover, we randomly select 5K candidates and we compare the true eigenvalues $\lambda'$ after performing a flip (i.e.\ doing a full eigen-decomposition) and our approximation $\tilde{\lambda'}$.
We found that the difference $|\lambda' - \tilde{\lambda'}|$ is negligible: several orders of magnitude smaller than the eigenvalues themselves.
The difference between the terms $|\sum_{r=1}^T \lambda_i^{'r} - \sum_{r=1}^T \tilde{\lambda}_i^{'r}|$ used in Lemma \ref{lemma:cor_loss} is similarly negligible.
Additionally, we compare the true singular values $\sigma_i(S)$ of the matrix $S$ and their respective upper bounds $d_{min}^{-1} |\sum_{r=1}^T\lambda_i^r| \ge \sigma_i(S)$ obtained from Lemma \ref{lemma:cor_loss}. The gap is different across graphs and it is relatively small overall. We plot all these quantities for all graphs in the appendix (Sec.\ \ref{sec:further}). 
These results together demonstrate the quality of our approximation.

\begin{figure*}[t!]
	\begin{minipage}{0.33\textwidth}
		\centering
		\includegraphics[width=0.99\linewidth]{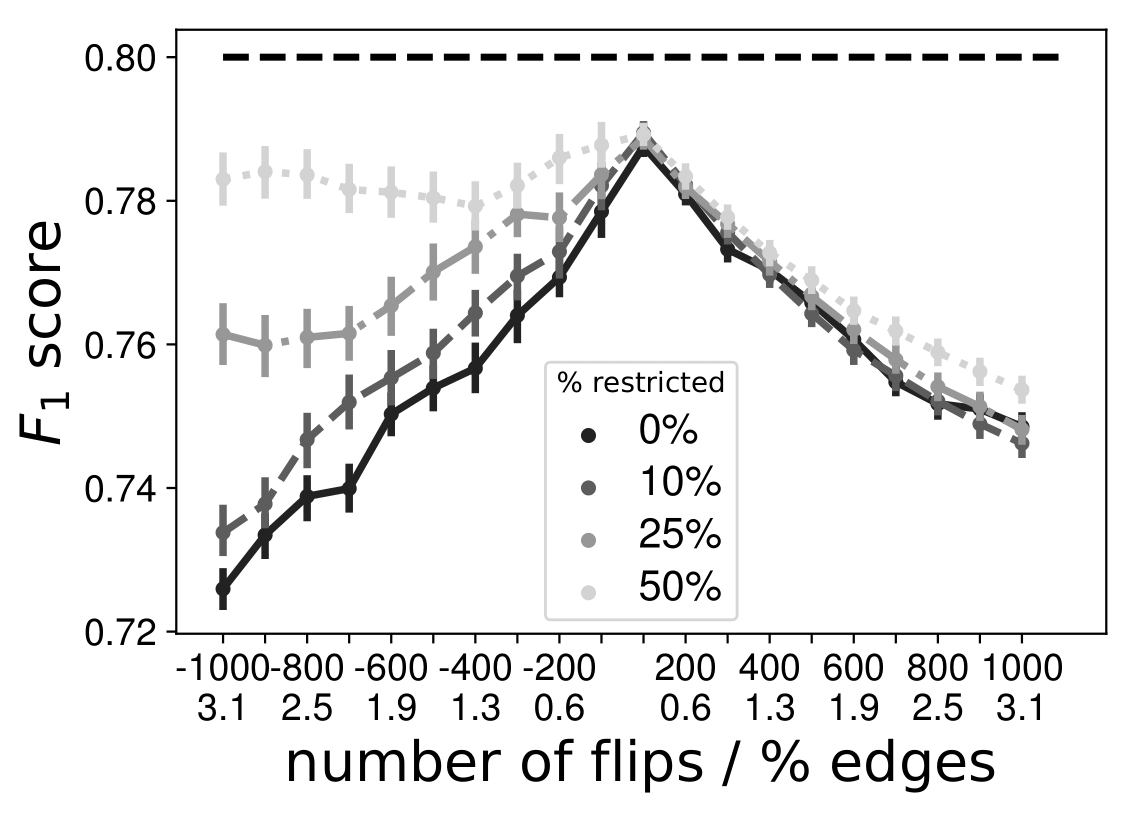}
		\caption{Cora: Classification performance for increasingly restricted attacks.}
		\label{fig_flips_hiding}
	\end{minipage}
	\begin{minipage}{0.66\textwidth}
		\centering
		\begin{subfigure}[h]{0.49\linewidth}
			\includegraphics[width=\textwidth]{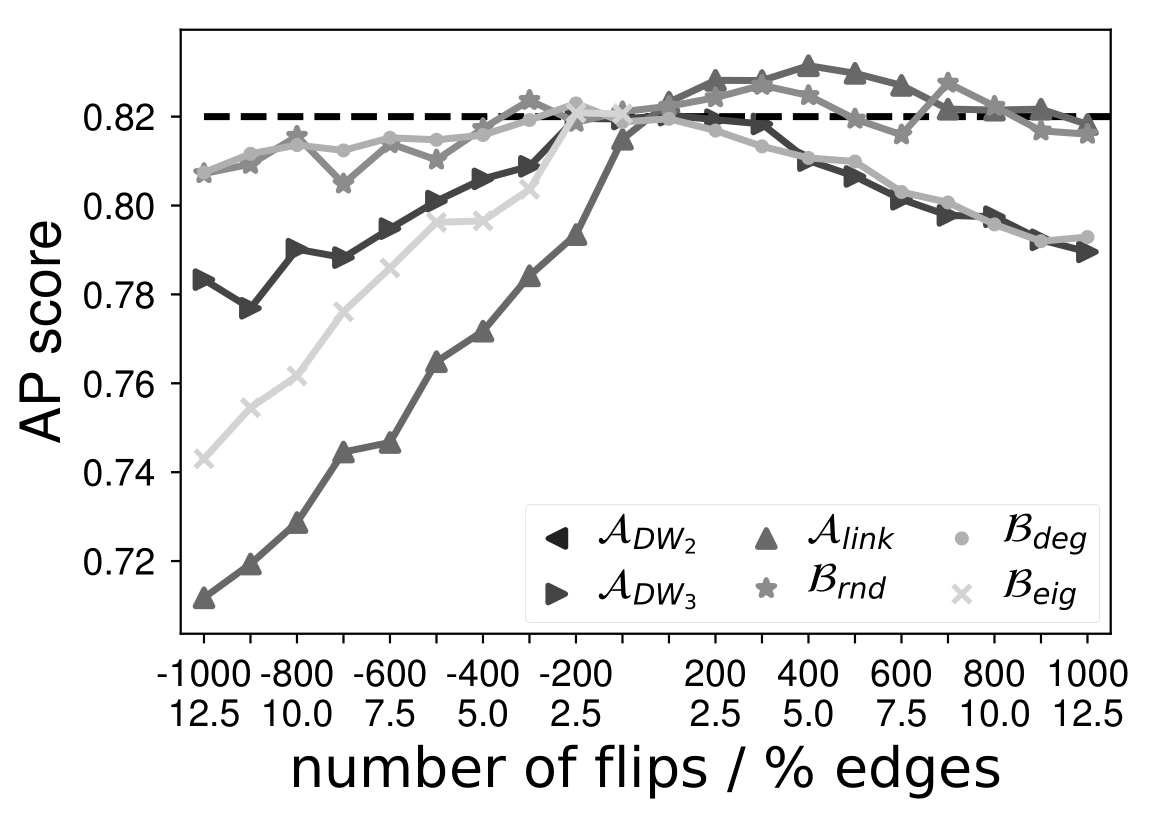}
			\caption{Cora}\label{fig:link_cora}
		\end{subfigure}
		\begin{subfigure}[h]{0.49\linewidth}
			\includegraphics[width=\textwidth]{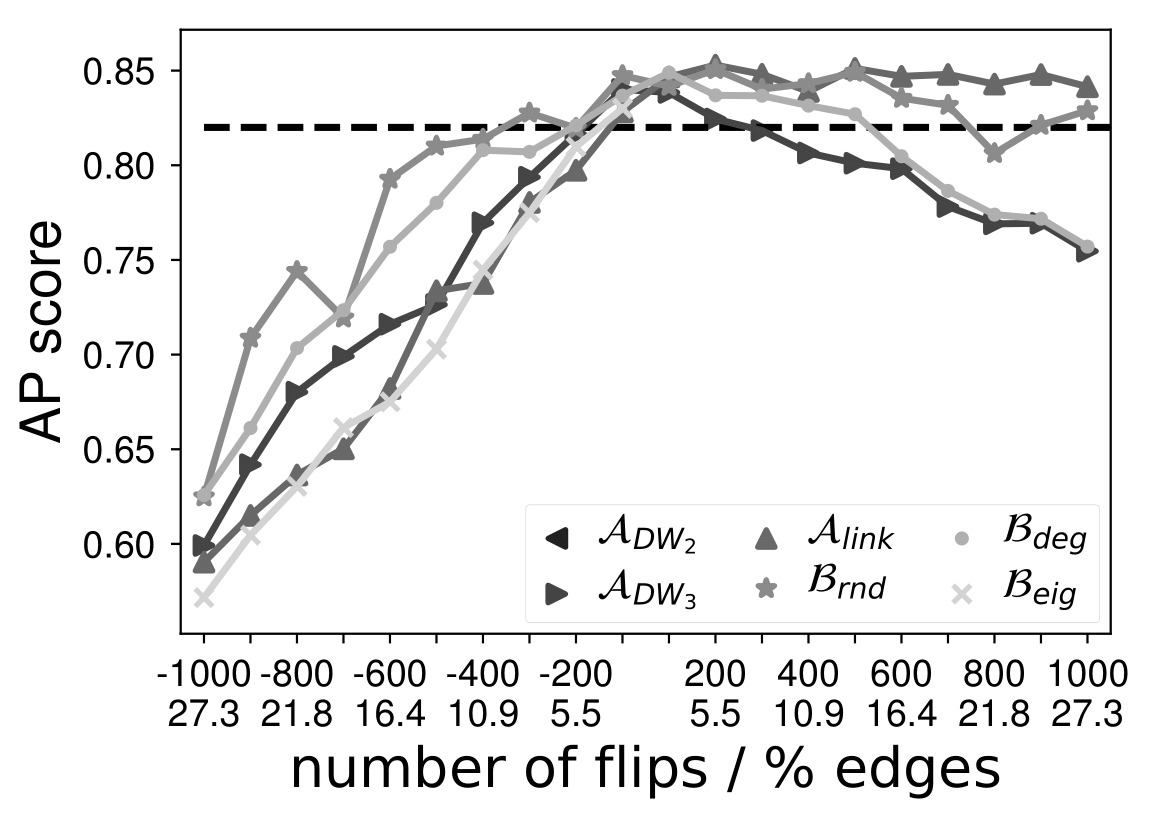}
			\caption{Citeseer}\label{fig:link_cite}
		\end{subfigure}
		\caption{Targeted attack on the link prediction task.}\label{fig:link}
	\end{minipage}
\end{figure*}
\subsection{General Attack}
\label{sec:gen_attack}
To better understand the attacks we investigate the effect of removing and adding edges separately. 
We select the top $f$ edges from the respective candidate sets according to our approximation of the loss function.
For adding edges, we also implemented an alternative add-by-remove strategy denoted as $\mathcal{A}_{abr}$. Here, we first add $cf$-many edges randomly sampled from the candidate set to the graph and subsequently remove $(c-1)f$-many of them (equals to only $f$ changes in total). This strategy performed better empirically.
Since the graph is undirected, for each $(i, j)$ we also flip $(j, i)$.

Fig.\ \ref{fig:attacks_first} answers question (Q2). Removed/added edges are denoted on the x-axis with negative/positive values respectively. 
On Fig.\ \ref{fig_flips_loss} we see that our strategies achieve a significantly higher loss compared to the baselines when removing edges.
To analyze the change in the embedding quality we consider the node classification task (i.e. using it as a proxy to evaluate quality; this is \emph{not} our targeted attack).
Interestingly, \attdeg is the strongest baseline w.r.t. to the loss, but this is not true for the downstream task.
As shown in Fig.\ \ref{fig_flips_f1} and \ref{polblogs}, our strategies significantly outperform the baselines. As expected, \attpert and $\mathcal{A}_{abr}$ perform better than \attgrad.
On Cora our attack can cause up to around 5\% \emph{more} damage compared to the strongest baseline. On PolBlogs, by adding only 6\% edges we can decrease the classification performance by more than 23\%, while being more robust to removing edges.

\textbf{Restricted attacks.}
In the real world attackers cannot attack any node, but rather only specific nodes under their control, which translates to restricting the candidate set.
To evaluate the restricted scenario, we first initialize the candidate sets as before, then we randomly denote a given percentage $p_r$ of nodes as restricted and discard every candidate that includes them.
As expected, the results in Fig.\ \ref{fig_flips_hiding} show that for increasingly restrictive sets with $p_r=10\%,25\%,50\%$, our attack is able to do less damage. However, we always outperform the baselines (not plotted), and even in the case when half of the nodes are restricted ($p_r=50\%$) we are still able to damage the embeddings. With this we can answer question (Q3): attacks are successful even when restricted.
\begin{figure}[b!]
	\centering
	\begin{subfigure}[h]{0.46\linewidth}
		\includegraphics[width=\textwidth]{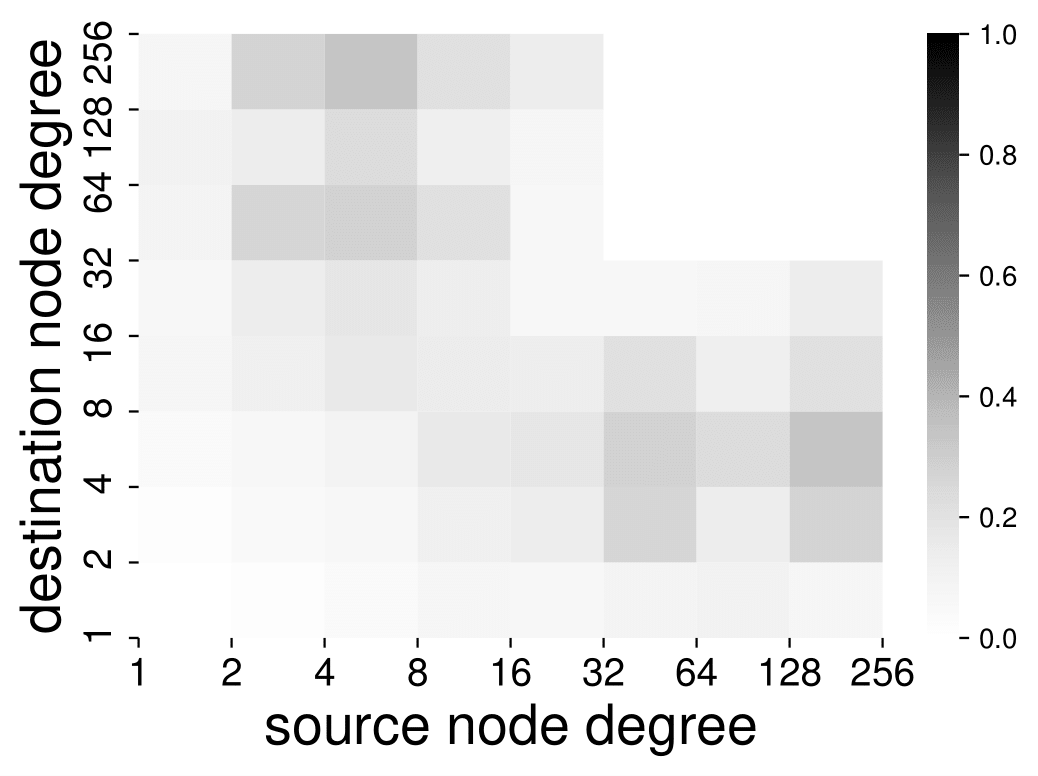}
		\caption{Degree centrality}
		\label{degrees}
	\end{subfigure}
	\begin{subfigure}[h]{0.49\linewidth}
		\includegraphics[width=\textwidth]{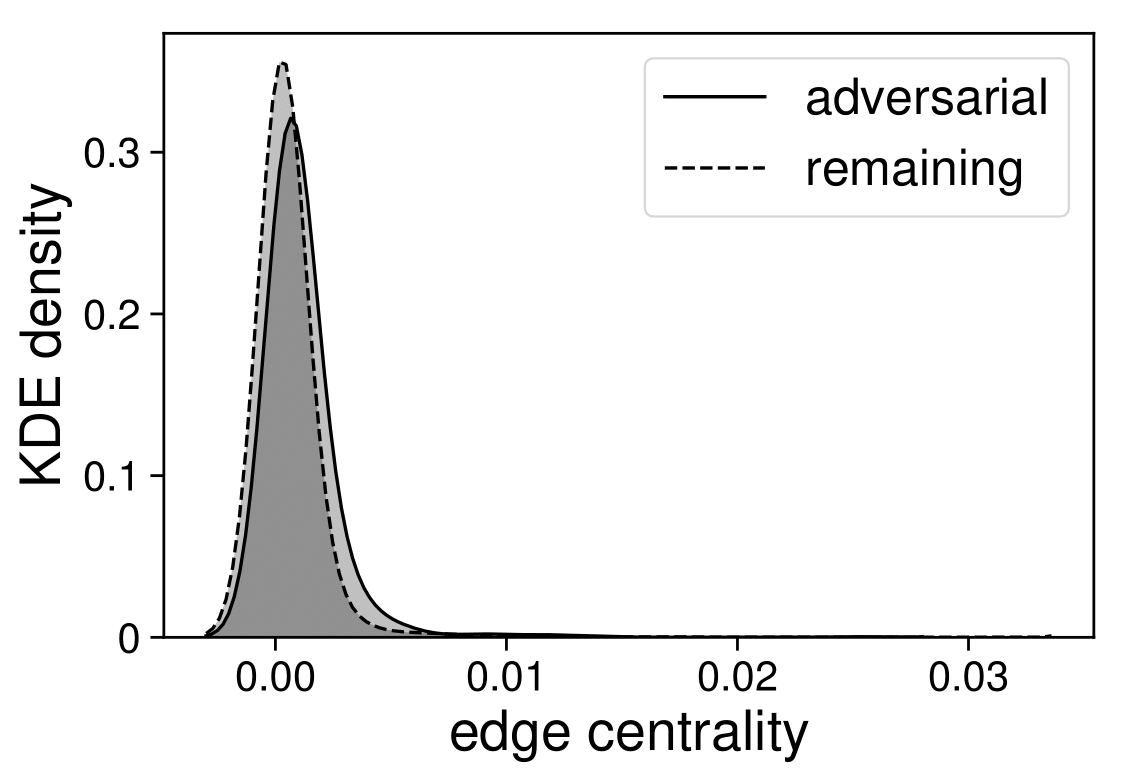}
		\caption{Edge centrality}\label{centrality}
	\end{subfigure}
	\caption{Analysis of the adversarial edges selected by \attpert on the Cora graph w.r.t. different centrality measures.}
\end{figure}

\begin{figure*}[t!]
	\begin{subfigure}[h]{0.245\linewidth}
		\centering
		\includegraphics[width=\textwidth]{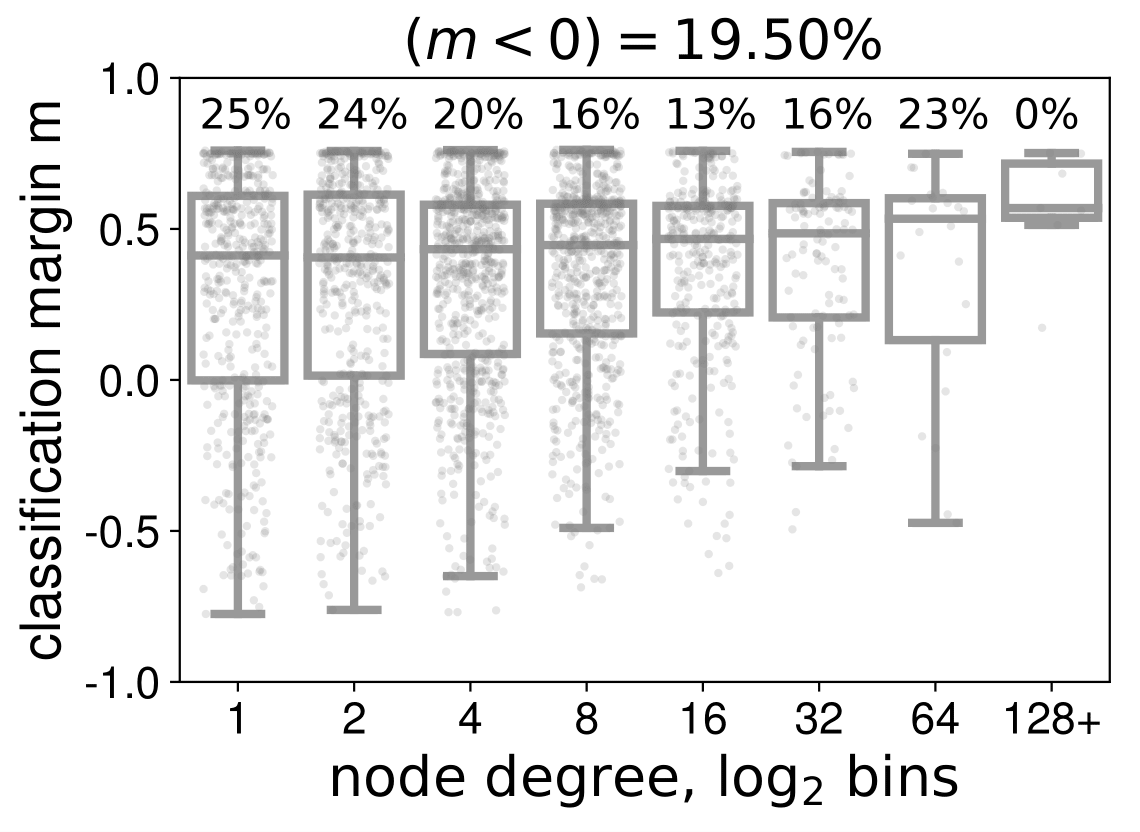}
		\caption{Before attack}
		\label{fig:targeted_before}
	\end{subfigure}
	\begin{subfigure}[h]{0.245\linewidth}
		\centering
		\includegraphics[width=\textwidth]{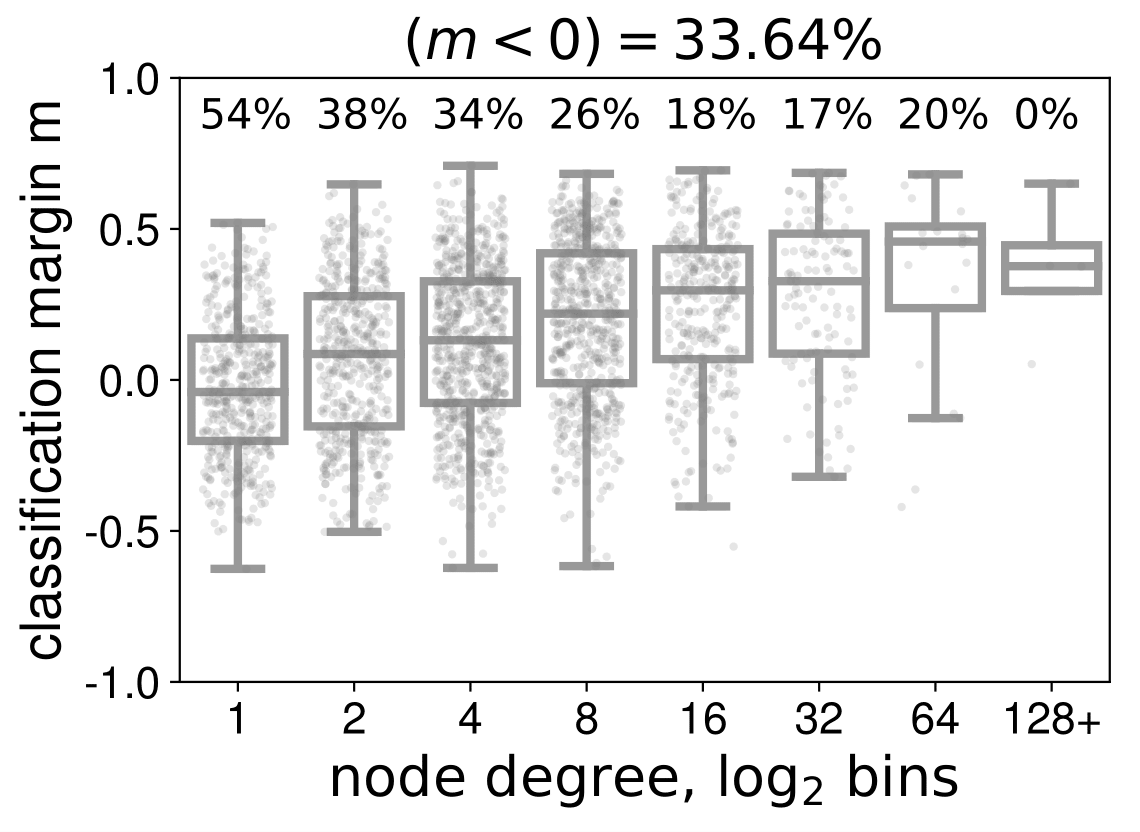}
		\caption{Baseline \attrand attack}
		\label{fig:targeted_rnd}
	\end{subfigure}
	\begin{subfigure}[h]{0.245\linewidth}
		\centering
		\includegraphics[width=\textwidth]{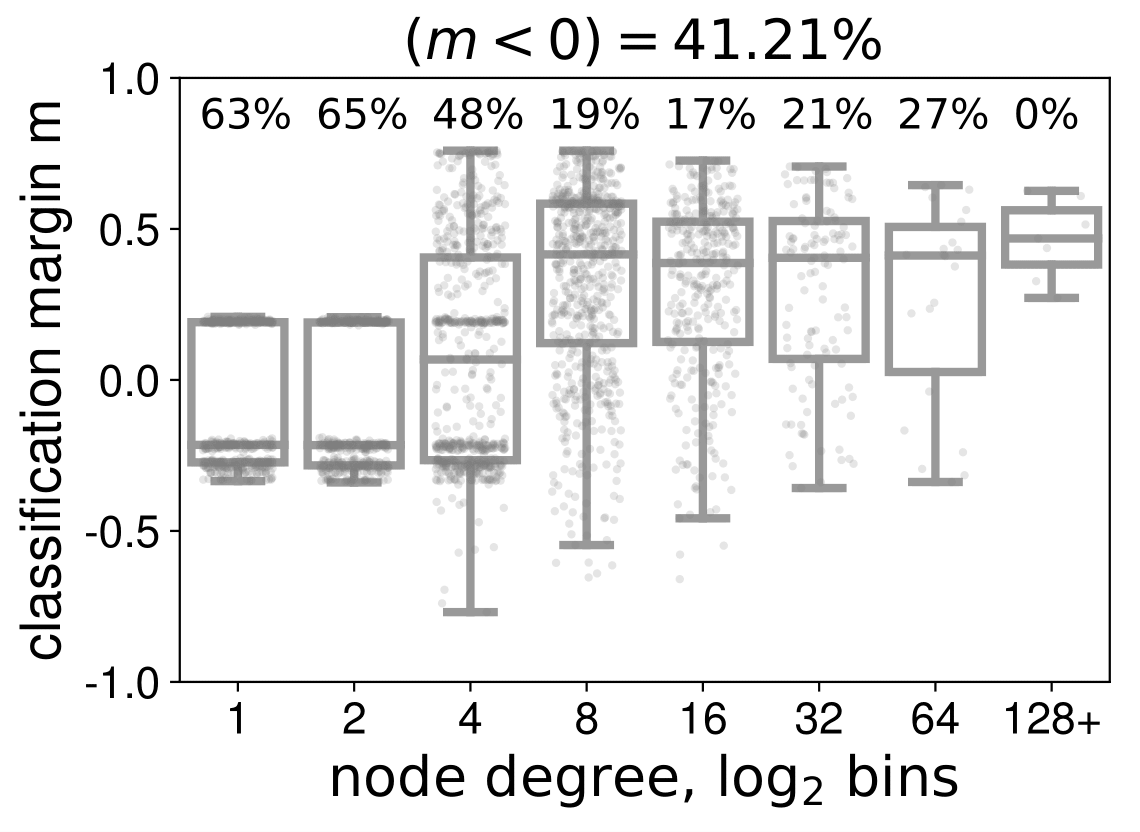}
		\caption{Baseline \attdeg attack}
		\label{fig:targeted_deg}
	\end{subfigure}
	\begin{subfigure}[h]{0.245\linewidth}
		\centering
		\includegraphics[width=\textwidth]{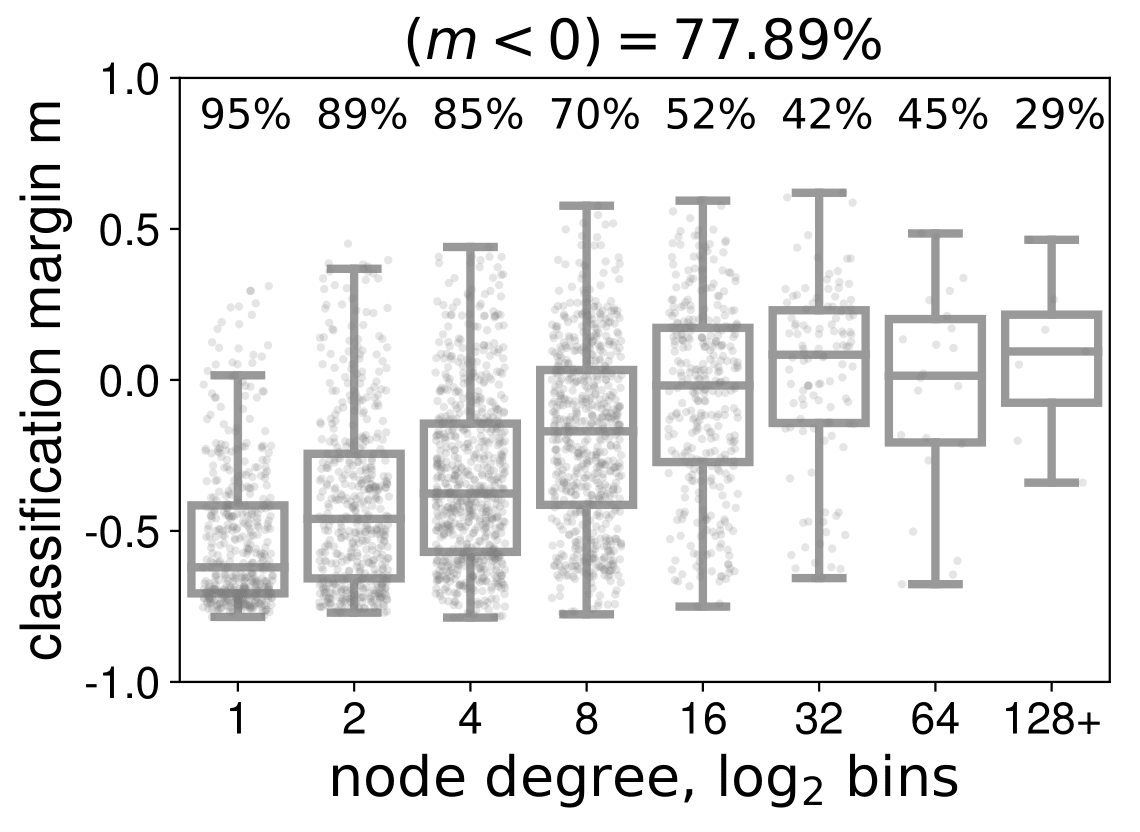}
		\caption{Our \atttar  attack}
		\label{fig:targeted_our}
	\end{subfigure}
	\caption{Margins for the clean and corrupted graphs for different attacks. Each dot represent one node binned logarithmically according to its degree. The number above each (box-) plot indicates the misclassification rate ($m<0$). Lower is better.
	}
	\label{fig:targeted}
\end{figure*}

\textbf{Analysis of the selected adversarial edges.}
A natural question to ask is what characterizes the adversarial edges that are selected by our attack, and whether its effectiveness can be explained by a simple heuristic such as attacking "important" edges (e.g. edges that have high centrality).
To answer this question we analyze the top 1000 edges selected by \attpert on the Cora dataset. 
In Fig.\ \ref{degrees} we analyze the adversarial edges in terms of node degrees.
Specifically, for each edge we consider the degree of its source node and its destination node and plot it on the x-axis and y-axis respectively. % 
The heatmap shows the number of adversarial edges divided by total number of edges for each degree (binned logarithmically). 
We see that low, medium and high degree nodes are all represented and therefore we conclude that we cannot distinguish between adversarial and non-adversarial edges based solely on their degrees.

In Fig.\ \ref{centrality} we plot the edge centrality distribution for the top 1000 adversarial edges and compare it with the edge centrality distribution of the remaining edges. We can see that there is no clear distinction. Both of these findings highlight the need for a principled method such as ours since using intuitive heuristics such as degree centrality or edge centrality cannot identify adversarial edges.

\subsection{Targeted Attack}
To obtain a better understanding of the performance we study the margin $m(t)$ (Sec.\ \ref{sec:targeted_class}) on Cora before and after the attack considering every node $t$ as a potential target. We allow a budget of only $(d_t+3)$ flips per each node (where $d_t$ is the degree of the target node $t$) ensuring that the degrees do not change noticeably after the attack.
Each dot in Fig.\ \ref{fig:targeted} represents one node grouped by its degree in the clean graph (logarithmic bins). We see that low-degree nodes are easier to misclassify ($m(t) < 0$), and that high degree nodes are more robust in general -- the baselines have $0\%$ success. Our method, however, can successfully attack even high degree nodes. 
In general, our attack is significantly more effective across all bins -- as shown by the numbers on top of each box -- with $77.89\%$ nodes successfully misclassified on average compared to e.g. only $33.64\%$ for \attrand.

For the link prediction task (Fig.\ \ref{fig:link}) we are similarly able to cause significant damage -- e.g. \attlnk achieves almost $10\%$ decrease in performance by flipping around $12.5\%$ of edges on Cora, significantly better than all other baselines. Here again, compared to adding edges, removing has a stronger effect. Overall, answering (Q5), both experiments confirm that our attacks hinder the various downstream tasks.
\subsection{Transferability}
\label{sec:transferability}
The question of whether attacks learned for one model generalize to other models is important since in practice the attacker might not know the model used by the defender. However, if transferability holds, such knowledge is not required.
To obtain the perturbed graph, we remove the top $f$ adversarial edges with \attpert. The same perturbed graph is used to learn embeddings using several other state-of-the-art approaches: DeepWalk (DW) with both the SVD and the SGNS loss, node2vec \cite{grover2016node2vec},
Spectral Embedding \cite{von2007tutorial}, Label Propagation \cite{zhu2002learning}, and GCN \cite{kipf2016semi}.

Table \ref{table_transfer} shows the change in node classification performance compared to the embeddings learned on the clean graph for each method respectively.
Answering (Q6), the results show that our attack generalizes: the adversarial edges have a noticeable impact on other models as well. We see that we can damage DeepWalk trained with the skip-gram objective with negative sampling (SGNS) showing that our factorization analysis via SVD is successful. We can even damage the performance of semi-supervised approaches such as Graph Convolutional Networks (GCN) and Label Propagation.

\begin{table}[h]
	\caption{Transferability: The change in $F_1$ score (in percent) compared to the clean/original graph. Lower is better.}
	\label{table_transfer}
	\centering
	\setlength{\tabcolsep}{2pt}
	\begin{small}
		\begin{tabular}{cc|cccccc}
&\multicolumn{1}{c|}{\multirow{2}{*}{Method}} & \multirow{2}{*}{\begin{tabular}[c]{@{}c@{}}DW\\ SVD\end{tabular}} & \multirow{2}{*}{\begin{tabular}[c]{@{}c@{}}DW\\ SGNS\end{tabular}} & \multirow{2}{*}{\begin{tabular}[c]{@{}c@{}}node-\\ 2vec\end{tabular}} & \multirow{2}{*}{\begin{tabular}[c]{@{}c@{}}Spect.\\ Embd\end{tabular}} & \multirow{2}{*}{\begin{tabular}[c]{@{}c@{}}Label\\ Prop.\end{tabular}} & \multirow{2}{*}{GCN} \\
& \multicolumn{1}{c|}{}&   &&   & & &                      \\ \hline
&			\multicolumn{1}{c|}{Cora} \\
\multirow{4}{*}{\rot{Our Approach}}		&	\multicolumn{1}{c|}{$f=250 (03.1\%)$}  & -3.59         & -3.97           & -2.04    & -2.11           & -5.78 & -3.34 \\
		&	\multicolumn{1}{c|}{$f=500 (06.3\%)$} & -5.22        & -4.71          & -3.48   & -4.57           & -8.95 & -2.33 \\ \cline{2-8}
		&	\multicolumn{1}{c|}{Citeseer} \\
		&	\multicolumn{1}{c|}{$f=250 (06.8\%)$ } & -7.59         & -5.73           & -6.45    & -3.58           & -4.99 & -2.21 \\
		&	\multicolumn{1}{c|}{$f=500 (13.6\%)$ }& -9.68        & -11.47          & -10.24   & -4.57           & -6.27 & -8.61 \\ 
	& \multicolumn{1}{c|}{}	\\ \hline
%		&	\multicolumn{1}{c|}{\atteig baseline} \\ \hline
		&\multicolumn{1}{c|}{	Cora }\\
\multirow{4}{*}{\rot{\atteig Baseline}}			&	\multicolumn{1}{c|}{$f=250 (03.1\%)$}  & -0.61         & -0.65           & -0.57    & -0.86          & -1.23 & -6.33 \\
	&	\multicolumn{1}{c|}{	$f=500 (06.3\%)$} & -0.71        & -1.22          & -0.64   & -0.51           & -2.69 & -0.64 \\ \cline{2-8}
		&	\multicolumn{1}{c|}{Citeseer} \\
		&	\multicolumn{1}{c|}{$f=250 (06.8\%)$}  & -0.40         & -1.16           & -0.26    & +0.11           & -1.08 & -0.70 \\
		&	\multicolumn{1}{c|}{$f=500 (13.6\%)$} & -2.15        & -2.33         & -1.01   & +0.38           & -3.15 & -1.40 \\
		\end{tabular}
	\end{small}
\end{table}

Compared to the transferability of the strongest baseline \atteig, shown in the lower section of Table \ref{table_transfer}, we can clearly see that our attack causes significantly more damage.
\section{Conclusion}
We demonstrated that node embeddings are vulnerable to adversarial attacks which can be efficiently computed and have a significant negative effect on downstream tasks such as node classification and link prediction. Furthermore, successfully poisoning the graph is possible with relatively small perturbations and under restriction. More importantly, our attacks generalize - the adversarial edges are transferable across different models.
Developing effective defenses against adversarial attacks as well as more comprehensive modelling of the attacker's knowledge are important directions for improving network representation learning.
\section*{Acknowledgments} 
This research was supported by the German Research Foundation, Emmy Noether grant GU 1409/2-1, and the German Federal Ministry of Education and Research (BMBF), grant no. 01IS18036B. The authors of this work take full responsibilities for its content.

\bibliography{paper}
\bibliographystyle{icml2019}

\newpage
\clearpage
\section{Appendix}
\subsection{Proofs and Derivations}
\begin{proof}
	\textbf{Theorem \ref{theo:naive_grad}.} Applying eigenvalue perturbation theory we obtain that $\lambda_p' = \lambda_p + u_p^T(\Delta \lm)u_p$ where $\lambda_p'$ is the eigenvalue of $\lm'$ based on $A'$ obtained after perturbing the graph.
	Using the fact that $\lambda_p=u_p^T\lm u_p$, and the fact that singular values are equal to the absolute value of the corresponding eigenvalues we obtain the desired result.
\end{proof}

\begin{proof}
	\textbf{Theorem \ref{theo:delta_lambda}.} Denote with $e_i$ the vector of all zeros and a single one at position $i$. Then, we have $\Delta A = \Delta w_{ij}(e_ie_j^T + e_je_i^T)$
	and $\Delta D = \Delta w_{ij}(e_ie_i^T + e_je_j^T)$. From eigenvalue perturbation theory \cite{stewart1990matrix}, we get:
	$
	\lambda'_y \approx \lambda_y + u_y^T (\Delta A - \lambda_y \Delta D) u_y
	$.
	Substituting $\Delta A$/$\Delta D$ concludes the proof.
\end{proof}

We include an immediate result to prove Theorem \ref{theo:delta_u}.
\begin{lemma}
\label{lem:delta_u_general}
Consider the generalized eigenvalue problem $Au=\lambda D u$ and suppose that we have the changes in the respective matrices/vectors: $\Delta A, \Delta D$ and $ \Delta\lambda$, then the change in the eigenvectors $\Delta u$ can be expressed as:
$$
\Delta u = -(A -\lambda D)^+ \bigg( \Delta A - \Delta \lambda D - \lambda \Delta D \bigg)u
$$
\end{lemma}

\begin{proof}
	\textbf{Theorem \ref{theo:delta_u}.} Let $\Delta A$ and $\Delta D$ be defined as in Theorem \ref{theo:delta_lambda} and let $\Delta \lambda$ be the change in the eigenvalues as computed in Theorem \ref{theo:delta_lambda}. Plugging these terms in Lemma \ref{lem:delta_u_general} and simplifying we obtain the result. 
\end{proof}

We include an intermediate result to prove Lemmas \ref{lemma:Seq} and \ref{lemma:cor_loss}.

\begin{lemma}
	\label{lemma:gen_eig_prob}
	 $\lambda$ is an eigenvalue of $D^{-1/2} A D^{-1/2}:=A_{norm}$ with eigenvector $\hat{u}=D^{1/2}u$ if and only if $\lambda$ and $u$ solve the generalized eigen-problem $Au=\lambda Du$. 
\end{lemma}

\begin{proof}
	\textbf{Lemma \ref{lemma:gen_eig_prob}.}
	We have $Az=\lambda Dz \implies (Q^{-1}AQ^{-T})(Q^Tz) = \lambda (Q^Tz)$ for any real symmetric $A$ and any positive definite $D$, where $D=QQ^T$ using the Cholesky factorization. Substituting the adjacency/degree matrix and using $Q=Q^T=D^{1/2}$ we obtain the result.
\end{proof}

\begin{proof}
	\textbf{Lemma \ref{lemma:Seq}.}
	$S$ is equal to a product of three matrices $S  = D^{-1/2} \big( \hat{U} \big(\sum_{r=1}^{T} \hat{\Lambda}^r \big) \hat{U}^T\big) D^{-1/2}$ 
	where  $\hat{U}\hat{\Lambda}\hat{U}^T = D^{-1/2} A D^{-1/2} =: A_{norm}$ is the eigenvalue decomposition of $A_{norm}$ (\citet{qiu2017network}).
	From Lemma \ref{lemma:gen_eig_prob} we have the fact that $\lambda$ is an eigenvalue of $D^{-1/2} A D^{-1/2}$ with eigenvector $\hat{u}=D^{1/2}u$ if and only if $\lambda$ and $u$ solve the generalized eigen-problem $Au=\lambda Du$. 
	Substituting $\hat{\Lambda}=\Lambda$ and $\hat{U}=D^{1/2}U$ in $S$, and use the fact that $D$ is diagonal.
\end{proof}

\begin{proof}	
	\textbf{Lemma \ref{lemma:cor_loss}.}
	Following 	\cite{qiu2017network}, the singular values of $S$ can be bounded by $\sigma_p(S) \le \frac{1}{d_{min}} \big\vert \sum_{r=1}^{T} (\hat{\mu}_{\pi(p)})^r \big\vert$ where $\mu$ are the (standard) eigenvalues of $A_{norm}$. Using Lemma \ref{lemma:gen_eig_prob}, the same bound applies using the generalized eigenvalues $\lambda_p$ of $A$. 
	Now using Theorem \ref{theo:delta_lambda}, we obtain $\tilde{\lambda'_p}$ an approximation of the p-th \emph{generalized} eigenvalue of $A'$. 
Plugging it into the singular value bound  we obtain:
	$\sigma_p(S) \le \frac{1}{d_{min}} \big\vert \sum_{r=1}^{T} (\tilde{\lambda}'_{\pi(p)})^r \big\vert$ which concludes the proof.
\end{proof}

Please note that the permutation $\pi$ does not need be computed/determined explicitly. In practice, for every $\tilde{\lambda}'_p$, we compute the term $\big\vert \sum_{r=1}^{T} (\tilde{\lambda}'_{p})^r \big\vert$. Afterwards, these terms are simply sorted.

%!TEX root = ../paper.tex
\begin{figure*}[t!]
	\centering
	\begin{subfigure}[h]{0.48\linewidth}
		\includegraphics[width=\textwidth]{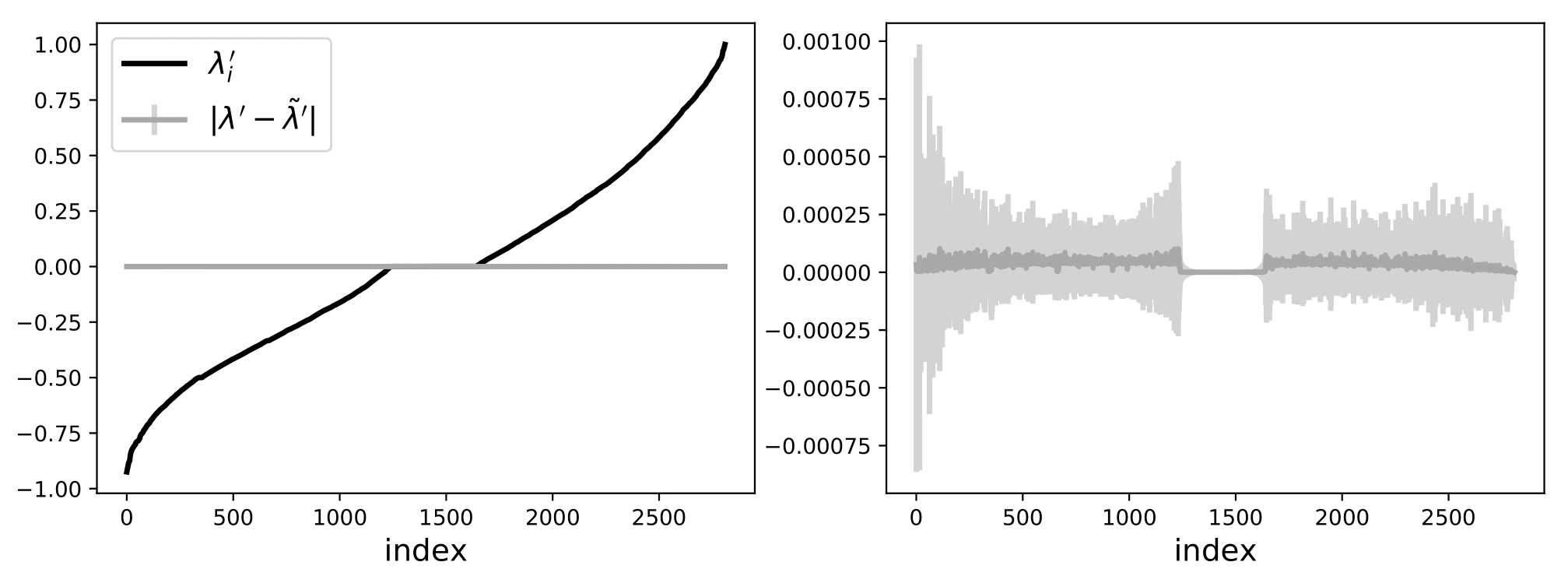}
		\caption{Eigenvalues: $|\lambda' - \tilde{\lambda'}|$}
		\label{fig:gaps_vals}
	\end{subfigure}
	\begin{subfigure}[h]{0.48\linewidth}
		\includegraphics[width=\textwidth]{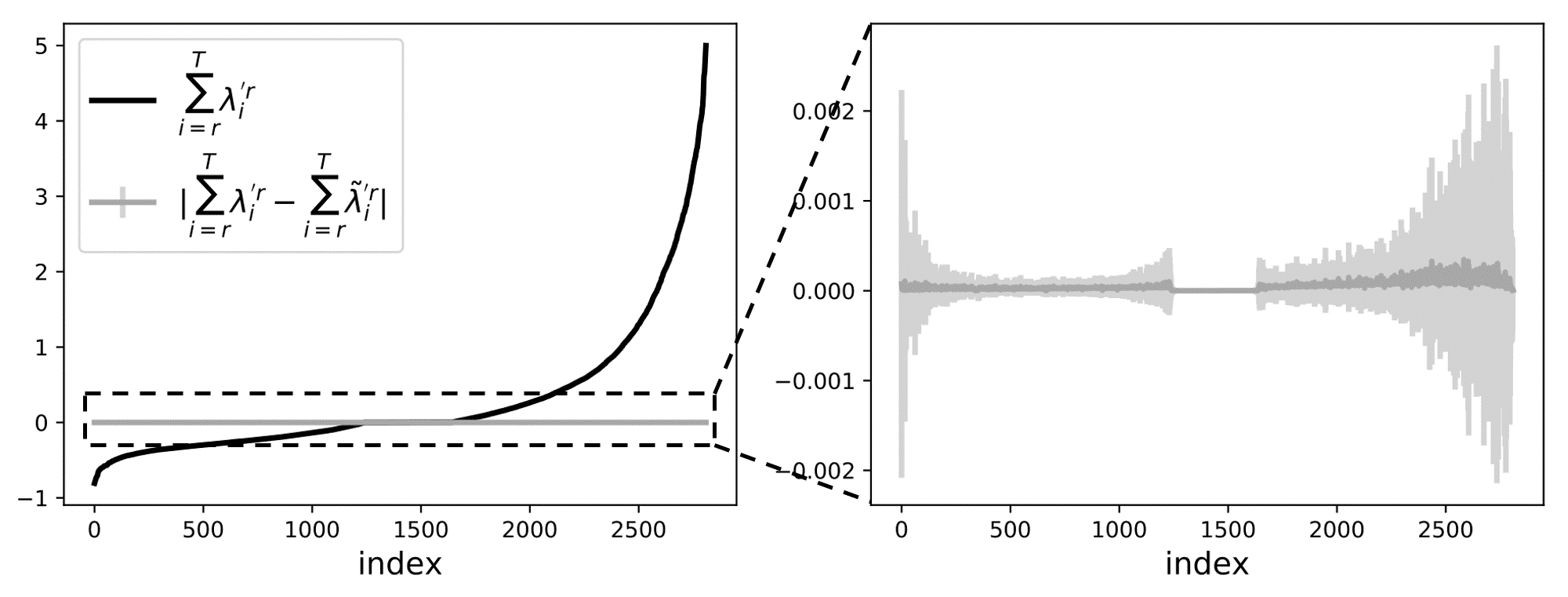}
		\caption{Sums of powers of eigenvalues: $\big|\sum_{r=1}^T \lambda_i^{'r} - \sum_{r=1}^T \tilde{\lambda}_i^{'r}\big|$.}
		\label{fig:gaps_sums}
	\end{subfigure}
	\caption{Comparison between the true eigenvalues $\lambda'$ after performing a flip (i.e.\ doing a full eigen-decomposition) and our approximation $\tilde{\lambda'}$. Since the difference is several orders of magnitude smaller than the eigenvalues (sums of powers of eigenvalues resp.) themselves, we show a "zoomed-in" view (note the difference in the scale on the y-axis). In each subplot on the right side we show the average absolute difference and the standard deviation across the 5K randomly selected flips. 
	}\label{fig:gaps}
\end{figure*}
\begin{figure*}[h!]
	\centering
	\begin{subfigure}[h]{0.24\linewidth}
		\includegraphics[width=\textwidth]{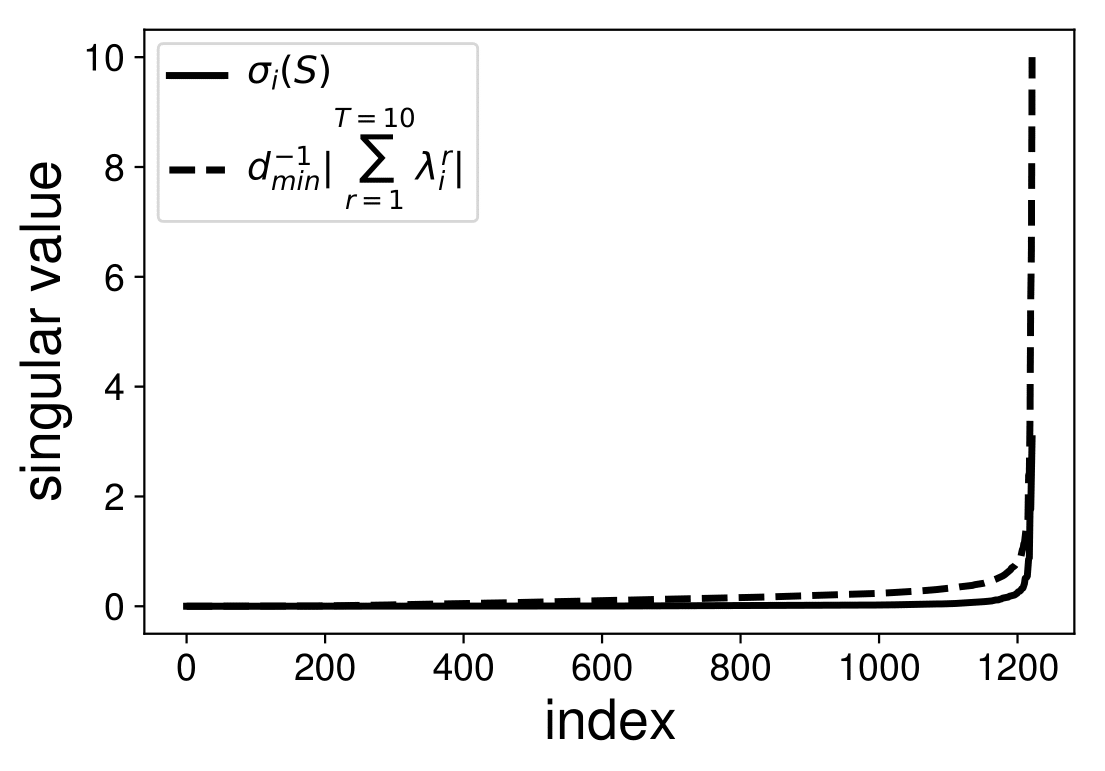}
		\caption{PolBlogs}
	\end{subfigure}
	\begin{subfigure}[h]{0.24\linewidth}
		\includegraphics[width=\textwidth]{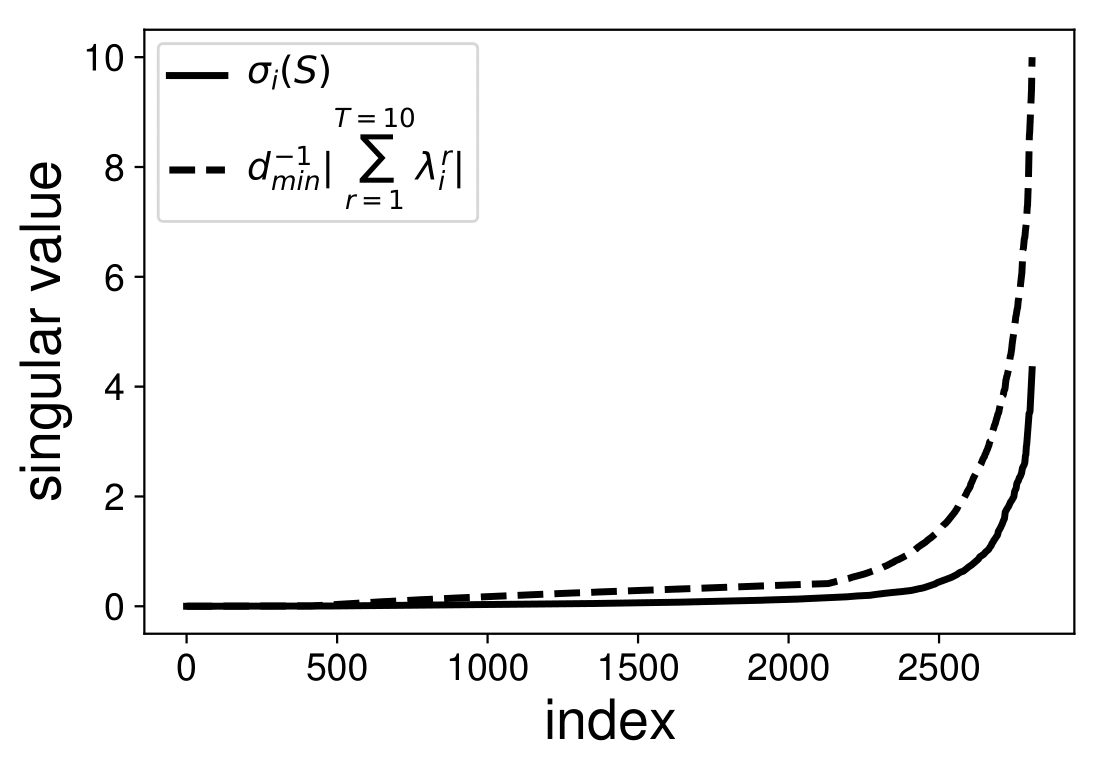}
		\caption{Cora}
	\end{subfigure}
	\begin{subfigure}[h]{0.24\linewidth}
		\includegraphics[width=\textwidth]{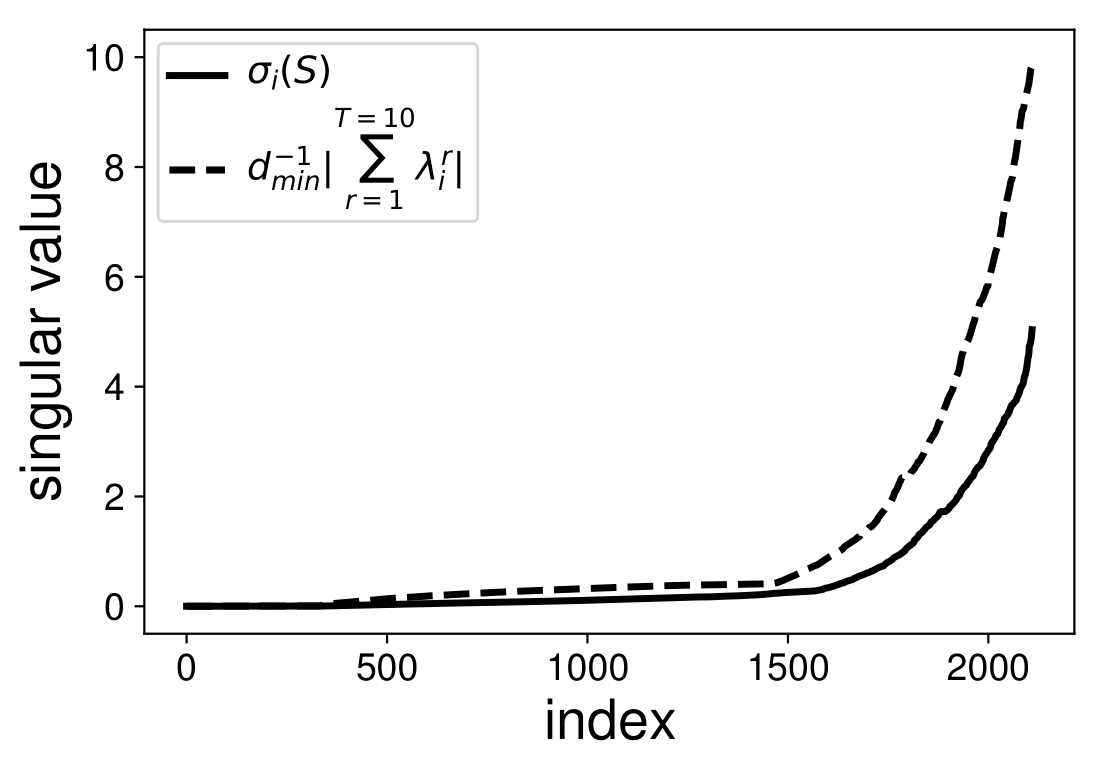}
		\caption{Citeseer}
	\end{subfigure}
	\caption{The singular value of $S$ and our upper bound $d_{min}^{-1} |\sum_{r=1}^T\lambda_i^r| \ge \sigma_i(S)$  for different graphs.}
	\label{fig:bounds}
\end{figure*}
\subsection{Analysis of Spectral Embedding Methods}
\label{sec:spectral}
\paragraph{Attacking spectral embedding.}
Finding the spectral embedding is equivalent to the following trace minimization problem:
\begin{equation}
	\label{loss_sc}
	\min_{Z \in \Set{R}^{|V| \times K}} Tr ( Z^T L_{xy} Z ) = \sum_{i=1}^{K} \lambda_i (L_{xy}) = \loss_{SC}
\end{equation}
subject to orthogonality constraints, where $L_{xy}$ is the graph Laplacian. The solution is obtained via the eigen-decomposition of $L$, with $Z^* = U_K$ where $U_K$ are the $K$-first eigen-vectors corresponding to the $K$-smallest eigenvalues $\lambda_i$.
The Laplacian is typically defined in three different ways: the unnormalized Laplacian $L=D-A$, the normalized random walk Laplacian $L_{rw} = D^{-1}L = I - D^{-1}A$ and the normalized symmetric Laplacian $L_{sym} = D^{-1/2} L D^{-1/2} = I - D^{-1/2} A D^{-1/2} = I - A_{norm}$, where $A, D, A_{norm}$ are defined as before.

\begin{lemma}[\cite{von2007tutorial}]
	\label{rw_sym}
	$\lambda$ is an eigenvalue of $L_{rw}$ with eigenvector $u$ if and only if $\lambda$ is an eigenvalue of $L_{sym}$ with eigenvector $w =D^{1/2}u$.
Furthermore, $\lambda$ is an eigenvalue of $L_{rw}$ with eigenvector $u$ if and only if $\lambda$ and $u$ solve the generalized eigen-problem $Lu=\lambda Du$.
\end{lemma}

From Lemma \ref{rw_sym} we see that we can attack both normalized versions of the graph Laplacian with a single attack strategy since they have the same eigenvalues. It also helps us to do that efficiently similar to our previous analysis (Theorem.~\ref{theo:delta_u}).

\begin{theorem}
	\label{delta_lambda_lap}
	Let $L_{rw}$ (or equivalently $L_{sym}$) be the initial graph Laplacian before performing a flip and $\lambda_y$ and $u_y$ be any eigenvalue and eigenvector of $L_{rw}$.
	The eigenvalue $\lambda'_y$ of  $L_{rw}'$ obtained after flipping a single edge $(i, j)$ 
	is
	\begin{equation}\label{est_loss_sc_norm}
	\lambda'_y \approx \lambda_y + \Delta w_{ij}( (u_{yi} - u_{yj})^2 - \lambda_y(u_{yi}^2 + u_{yj}^2 ))
	\end{equation}
	where $u_{yi}$ is the $i$-th entry of the vector $u_y$.
\end{theorem}
\begin{proof}
	From Lemma \ref{rw_sym} we can estimate the change in $L_{rw}$ (or equivalently $L_{sym}$) by estimating the eigenvalues solving the generalized eigen-problem $Lu=\lambda Du$.
	Let $\Delta L = L' - L$ be the change in the unnormalized graph Laplacian after performing a single edge flip $(i, j)$ and $\Delta D$ be the corresponding change in the degree matrix.
	Let $e_i$ be defined as before. Then $\Delta L = (1-2A_{ij})(e_i - e_j)(e_i - e_j)^T$ and $\Delta D = (1-2A_{ij})(e_ie_i^T + e_je_j^T)$. Based on the theory of eigenvalue perturbation we have $\lambda_y' \approx \lambda_y + u_y^T(\Delta L - \lambda_y\Delta D)u_y$.
	Finally, we substitute $\Delta L$ and $\Delta D$. 
\end{proof}

Using now Theorem \ref{delta_lambda_lap} and Eq.~\ref{loss_sc} we finally estimate the loss of the spectral embedding after flipping an edge  $\loss_{SC}(L_{rw}', Z) \approx \sum_{p=1}^{K} \lambda_p'
$. Note that here we are summing over the $K$-first \emph{smallest} eigenvalues.
 We see that spectral embedding and the random walk based approaches are indeed very similar. 

\begin{theorem}
	\label{theo:sc_unnorm}
	Let $L$ be the initial unnormalized graph Laplacian before performing a flip and $\lambda_y$ and $u_y$ be any eigenvalue and eigenvector of $L$. 	
	The eigenvalue $\lambda'_y$ of  $L'$ obtained after flipping a single edge $(i, j)$ 
	can be approximated by:
	\begin{equation}\label{est_loss_sc_unnorm}
	\lambda'_y \approx \lambda_y - (1-2A_{ij})(u_{yi} - u_{yj})^2
	\end{equation}
\end{theorem}
\begin{proof}
	Let $\Delta A = A' - A$ be the change in the adjacency matrix after performing a single edge flip $(i, j)$ and $\Delta D$ be the corresponding change in the degree matrix. Let $e_i$ be defined as before. 
	Then $\Delta L = L' - L = (D+\Delta D)-(A+\Delta A) - (D-A) = \Delta D - \Delta A = 
	(1-2A_{ij}) (e_ie_i^T + e_je_j^T - (e_ie_j^T + e_je_i^T))
	$. Based on the theory of eigenvalue perturbation we have $\lambda_y' \approx \lambda_y + u_y^T(\Delta L)u_y$. Substituting $\Delta L$ and re-arranging we get the above results. 
\end{proof}
\subsection{Approximation Quality}
\label{sec:further}
\textbf{Approximation quality of the eigenvalues.} 
We randomly select 5K candidate edge flips (Cora) and we compare the true eigenvalues $\lambda'$ after performing a flip (i.e.\ doing a full eigen-decomposition) and our approximation $\tilde{\lambda'}$ obtained from Theorem \ref{theo:delta_lambda}.
We can see in Fig.\ \ref{fig:gaps_vals} that the average absolute difference $|\lambda' - \tilde{\lambda'}|$ across the 5K randomly selected flips and the standard deviation are negligible: several orders of magnitude smaller than the eigenvalues themselves.
The difference between the terms $|\sum_{r=1}^T \lambda_i^{'r} - \sum_{r=1}^T \tilde{\lambda}_i^{'r}|$ used in Lemma \ref{lemma:cor_loss} is similarly negligible as shown in Fig. \ref{fig:gaps_sums}.

\textbf{Upper bound on the singular values.}
Lemma \ref{lemma:cor_loss} shows that $\mathcal{L}_{DW3}$ is an upper bound on $\mathcal{L}_{DW1}$ (excluding the elementwise logarithm). For a better understanding of the tightness of the bound we visualize the true singular values $\sigma_i(S)$ of the matrix $S$  and their respective upper bounds $d_{min}^{-1} |\sum_{r=1}^T\lambda_i^r| \ge \sigma_i(S)$ obtained by applying Lemma \ref{lemma:cor_loss} for all datasets. As we can see in Fig.\ \ref{fig:bounds}, the gap is different across the different graphs and it is relatively small overall. 

These results together (Fig. \ref{fig:gaps} and Fig. \ref{fig:bounds}) demonstrate that we have obtained a good approximation of both the eigenvalues and the singular values, leading to a good overall approximation of the loss.

\end{document}